%% file: neurips_2025.tex
\documentclass{article}

\PassOptionsToPackage{numbers, compress}{natbib}

\usepackage[preprint]{neurips_2025}

\usepackage[utf8]{inputenc} %
\usepackage[T1]{fontenc}    %
\usepackage{hyperref}       %
\usepackage{url}            %
\usepackage{booktabs}       %
\usepackage{amssymb}
\usepackage{amsmath}
\usepackage{amsfonts}       %
\usepackage{amsthm}
\usepackage{nicefrac}       %
\usepackage{microtype}      %
\usepackage{xcolor}         %
\usepackage{tikz}
\usepackage{pgfplots}

\usepackage[normalem]{ulem}

\usepackage{enumerate}
\usepackage{yfonts}
\usepackage[textwidth=65mm]{todonotes}
\usepackage{graphicx}

\usepackage{subcaption}
\usepackage{wrapfig}
\usepackage{psfrag}
\usepackage{epsfig}

\usepackage{cancel}
\usepackage{cleveref}

\usepackage{slashed}
\usepackage{multirow}
\usepackage{pifont}
\usepackage{algorithm}
\usepackage{algpseudocode}

\usepackage{enumerate}
\usepackage{enumitem}

\usepackage{bm}
\usepackage[all]{xy}

\input{thm-env}

\input{math-commands}

\input{more-commands}

\title{AdS-GNN - a Conformally Equivariant Graph Neural Network}

\author{
\makebox[\textwidth+2cm][c]{%
  \normalfont
  \hspace{-7em}\begin{tabular}[t]{c}
    \textbf{Maksim Zhdanov}\thanks{Equal contribution.} \\
    AMLab, University of Amsterdam\\
    \texttt{m.zhdanov@uva.nl}
  \end{tabular}
  \begin{tabular}[t]{c}
    \textbf{Nabil Iqbal}$^\ast$ \\
    Department of Mathematical Sciences, Durham University \\
    AMLab, University of Amsterdam\\
    \texttt{nabil.iqbal@durham.ac.uk}
  \end{tabular}%
}
  \AND
  Erik Bekkers \\
  AMLab, University of Amsterdam\\
  \texttt{e.j.bekkers@uva.nl} \\
  \And
  Patrick Forr\'e\\
  AMLab, AI4Science Lab, University of Amsterdam\\
  \texttt{p.d.forre@uva.nl}
}

\begin{document}

\maketitle

\begin{abstract}
  Conformal symmetries, i.e.\ coordinate transformations that preserve angles, play a key role in many fields, including physics, mathematics, computer vision and (geometric) machine learning. Here we build a neural network that is equivariant under general conformal transformations. To achieve this, we lift data from flat Euclidean space to Anti de Sitter (AdS) space. This allows us to exploit a known correspondence between conformal transformations of flat space and isometric transformations on the AdS space. We then build upon the fact that such isometric transformations have been extensively studied on general geometries in the geometric deep learning literature. We employ message-passing layers conditioned on the proper distance, yielding a computationally efficient framework. We validate our model on tasks from computer vision and statistical physics, demonstrating strong performance, improved generalization capacities, and the ability to extract conformal data such as scaling dimensions from the trained network. 
\end{abstract}

\section{Introduction}
The notion of {\it symmetry} is a key tool both in our understanding of nature and for the construction of machine learning systems that perceive nature. The construction of {\it equivariant} neural network architectures that encode specific symmetries has powerful advantages both conceptual and computational. In particular, much work has been dedicated to building networks that are equivariant under familiar symmetries such as rotations and translations. 

In this work, we will study the symmetry group of {\it conformal transformations} i.e. the set of transformations on $\mathbb{R}^d$ that preserve angles. This includes translations, rotations, reflections, scalings and so called special conformal transformations. Scale transformations -- i.e. rigid rescalings of the whole system -- are an extremely important subgroup of the conformal group. Conformal and scale invariance play a central role in many diverse fields. To give some examples: biological visual systems seem to exhibit insensitivity to scale \cite{logothetis1995shape, han2020scale}. Physical systems undergoing a second-order phase transition have fluctuations at all scales; they are generally conformally invariant at the critical point \cite{cardy1996scaling}, and can usefully be described by {\it conformal field theories} \cite{DiFrancesco:1997nk}. Indeed in physics most systems exhibiting scale invariance also exhibit conformal invariance ``for free'' \cite{Polchinski:1987dy}, \cite{Nakayama:2013is}. Diverse applications also exist in computational geometry and computer vision (see e.g.  \cite{sharon20062d,lei2023computational}) %

It is reasonable to believe that a neural network that is equivariant under conformal transformations will be naturally insensitive to scale, instead focusing on robust properties of shape and form, lending to many possible applications. In this work, we construct such a conformally equivariant network. Our approach acts on point clouds and lifts the data into an auxiliary higher dimensional space called Anti de Sitter ($\AdS$) space. As outlined below, this approach is inspired by ideas in conformal field theory in theoretical physics. We validate our construction on some tasks drawn from computer vision and statistical physics.  

\section{Previous work}

Construction of equivariant neural networks under general symmetry groups began with \cite{cohen2016group} and has since evolved into a rich, well-studied field (see, e.g., \cite{bronstein2021geometric, weiler2023EquivariantAndCoordinateIndependentCNNs} for reviews). Advancements include treatments of isometric transformations on general Riemannian manifolds \cite{weiler2023EquivariantAndCoordinateIndependentCNNs} and extensions to semi-Riemannian manifolds \cite{zhdanov2023implicit, zhdanov2024cliffordsteerable}. The significant benefits of incorporating inherent equivariance, particularly for point cloud data and even at large computational scales, are increasingly demonstrated \cite{vadgama2025utility, brehmer2024does}, highlighting advantages in performance and efficiency. Lie group theory has been employed to develop equivariant networks for broad classes of transformations, such as general affine transformations \cite{mironenco2024lie}, or to leverage specific algebraic structures like adjoint actions on Lie algebras \cite{lin2024lie}. However, despite progress in handling Euclidean symmetries, scale equivariance, or even these more general Lie groups, these methods do not address the full conformal group, which uniquely includes non-affine special conformal transformations.

The pursuit of scale equivariance, a critical component of conformal symmetry, is perhaps the closest related area to our work on full conformal equivariance, with various approaches developed from specialized convolutional architectures \cite{Bekkers2020B-Spline, sosnovik2019scale} to techniques like Fourier layers for robust scale handling \cite{rahman2023truly}. Our use of $\AdS$ space connects conceptually to \textit{scale space} theories \cite{witkin1987scale, worrall2019deep}, where the extra dimension corresponds to scale. However, a crucial distinction is that the geometry of $\AdS$ space enforces equivariance under the larger group of all conformal transformations by construction, not just scale or isometric transformations. This approach also finds resonance in physics, for instance, with recent explorations of using neural networks to model aspects of conformal field theory \cite{Halverson:2024axc}.

\input{figures/cg_main_fig}

\section{Conformal symmetry and Anti de Sitter space}
We briefly review conformal symmetry and the geometry of $\AdS$ space before describing our neural network. 

\subsection{Conformal transformations}

A detailed and self-contained account of conformal transformations is given in Appendix \ref{app:conf}; here we present a brief review. Formally, a \emph{global conformal transformation} of the Euclidean space $\mathbb{R}^d$ is 
an injective smooth map
$\varphi:\, \mathbb{R}^d \setminus \{x_\varphi\} \to \mathbb{R}^d$, $x \mapsto \varphi(x)$, defined on $\mathbb{R}^d$ except on a possible point $x_\varphi$ \footnote{By allowing $\varphi$ to not be defined on a certain point $x_\varphi \in \mathbb{R}^d$, we effectively allow $\varphi$ to map $x_\varphi$ to the ``points at infinity'' $\infty$ in the conformal compactification $\mathbb{S}^d$ of $\mathbb{R}^d$. In fact, every \emph{global} conformal transformation of $\mathbb{R}^d$  uniquely extends to an angle-preserving diffeomorphism of $\mathbb{S}^d$ for $d \ge 2$, see \cite{Scho08} Thm. 2.6-2.11, or \ref{thm:conf-grp} and \ref{thm:conf-grp-ext}. This is why we introduce the definition with $x_{\phi}$ in this way.} such that local angles are always preserved, i.e.\ for all $x \in \mathbb{R}^d\setminus \{x_\varphi\}$ and $v_1,v_2 \in \mathbb{R}^d \setminus \{0\}$ we require\footnote{An equivalent, but more general and abstract, definition is provided in \Cref{def:conf-map} and \Cref{lem:conf-angle-pres}.}:
\begin{align}
 \frac{\langle \varphi'(x) v_1,\varphi'(x) v_2 \rangle}{\|\varphi'(x) v_1\| \cdot \|\varphi'(x) v_2\|} 
 =: \cos(\measuredangle(\varphi'(x) v_1,\varphi'(x) v_2) ) \overset{!}{=}
 \cos(\measuredangle(v_1,v_2)) := \frac{\langle v_1,v_2 \rangle}{\|v_1\| \cdot \|v_2\|},
\end{align}
where $\langle.,.\rangle$ and $\|.\|$ denote the standard Euclidean scalar product and norm, resp., and $\varphi'(x)$ the Jacobian matrix of $\varphi$ at $x$.  Note, that, in contrast to \emph{isometric transformations}, we do not require that the distances/norms are preserved. The \emph{group of all global conformal transformations} of $\mathbb{R}^d$ is denoted by $\GConf(\mathbb{R}^d)$.
\Cref{thm:conf-grp} (see \cite{Scho08} Thm. 2.6-2.11) shows that for $d \ge 2$ the global conformal group $\GConf(\mathbb{R}^d)$ is isomorphic to the \emph{projective orthogonal group}: %
$\POr(d+1,1):=\Or(d+1,1)/\{\pm 1\}$, where the role of the quotient is described in \Cref{app:conf} in \Cref{prp:PO-conf}.
It is instructive to consider the action of the group on points in terms of separated parameters. 
In particular, a general element $G$ of the group can be written in terms of parameters $(\lambda, t, b, M) \in \mathbb{R}_{>0} \times \mathbb{R}^{d} \times \mathbb{R}^{d} \times \Or(d)$ and acts on a point $x \in \mathbb{R}^{d}$ through the composition of maps defined by:
\vspace{-2pt}
\begin{align}
&x' = x + t \qquad &&x'= M x \qquad &&x' = \lambda x \qquad &&\frac{x'}{\|x'\|^2} = \frac{x}{\|x\|^2} - b \label{confgroup}, \\
&\hspace{-6pt}\includegraphics[width=2cm]{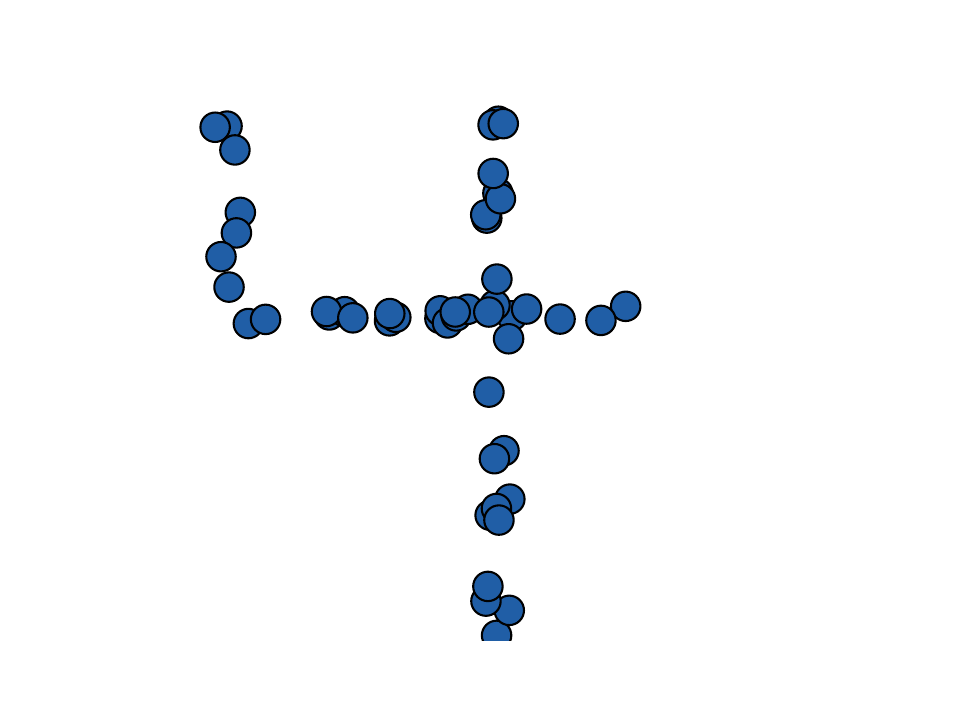} \qquad 
&&\hspace{-10pt}\includegraphics[width=2cm]{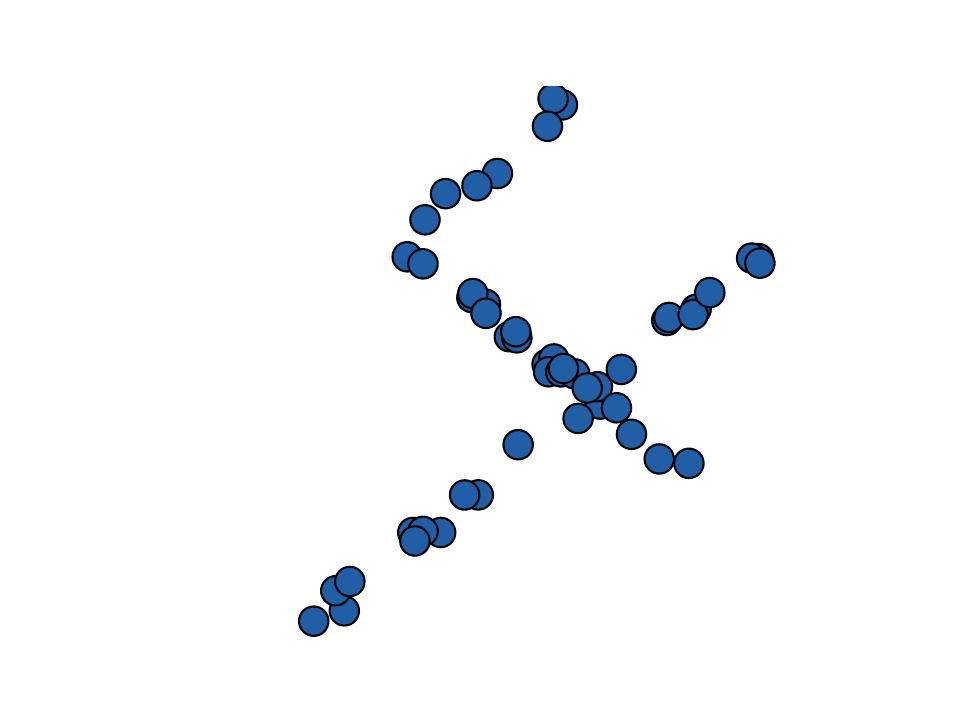} \qquad 
&&\hspace{-16pt}\includegraphics[width=2cm]{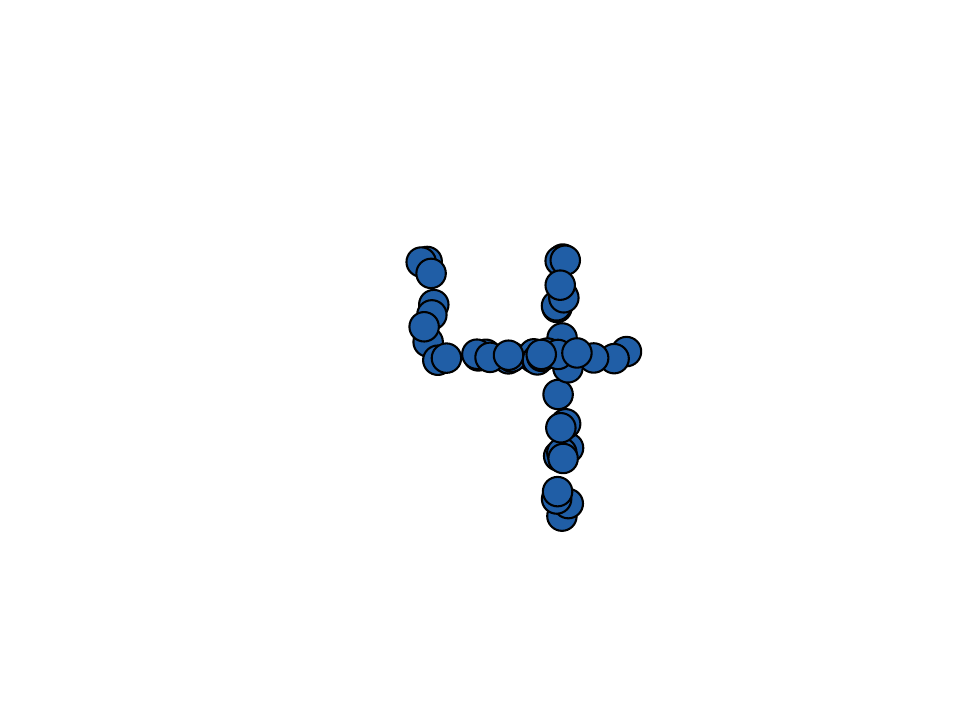} \qquad 
&&\hspace{10pt}\includegraphics[width=2cm]{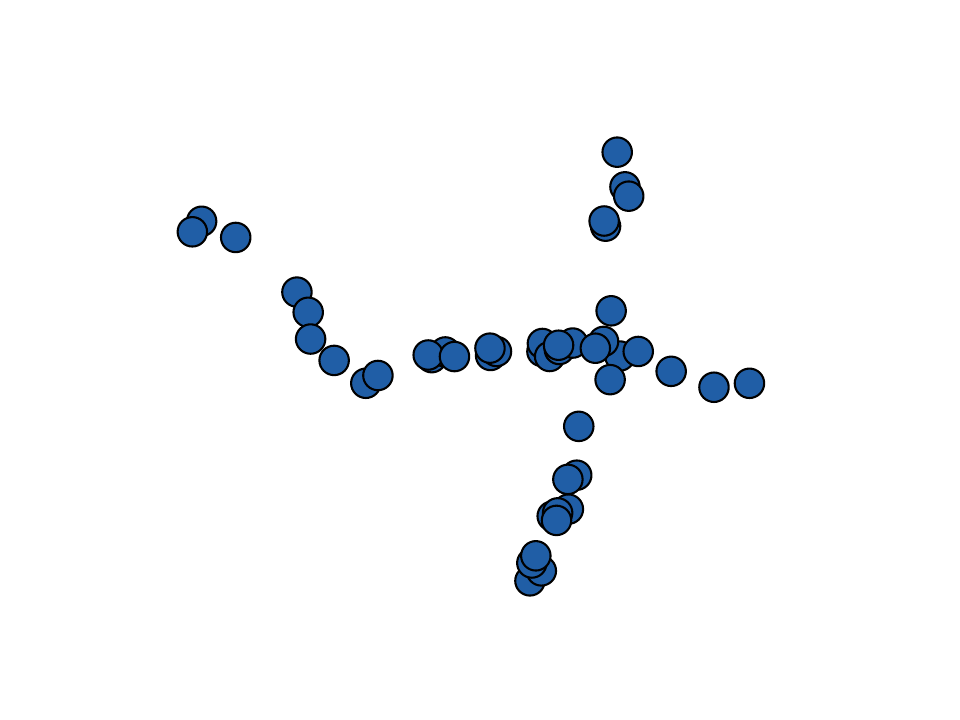} \notag \\
&\hspace{-2pt}\text{translations} &&\hspace{1pt}\text{rotations} &&\text{scalings} &&\hspace{-20pt}\text{special conformal transformations} \notag
\end{align}

resulting in a transformed point $x' = G x$. 
These corner cases are treated in detail in \Cref{eg:aff-conf-diff} and \ref{eg:inversion}.
These $(d+1)(d+2)/2$ parameters then assemble into an element of $\Or(d+1,1)$, see e.g. \cite{DiFrancesco:1997nk} for a review.   

Above we have explained how points $x \in \mathbb{R}^{d}$ transform under the global conformal group. In a typical application we will often be dealing with {\it conformal fields} $\phi(x)$ defined on this space. The transformation of these fields is governed by the representation theory of the global conformal group; elementary reviews can be found in \cite{Simmons-Duffin:2016gjk,DiFrancesco:1997nk}. 

This representation theory is non-trivial and we do not review it here, except to state that a privileged role is played by a basis of fields called {\it conformal primaries}, which we denote by $\sO(x)$. Associated with each primary field is a number $\Delta_{\sO}$ called the {\it conformal dimension} which characterizes the field in question\footnote{It may be helpful to consider fields transforming under the rotation group $\SOr(3)$: there are scalar fields, spinor fields, vector fields, etc. which are characterized by a discrete parameter called the {\it spin} $s$ which takes discrete values $s \in \frac{\mathbb{Z}}{2}$, i.e. $s_{\mathrm{scalar}} = 0, s_{\mathrm{vector}} = 1$, etc. $\Delta$ can very loosely be thought of as the analogue of ``spin'' for scale transformations. The fact that it is now a continuous variable is related to the fact that the conformal group is non-compact.}. 
Conformal primaries are distinguished by the fact that they transform as simply as possible under the group, and in particular under scaling of the coordinate they transform by a multiplicative factor: 
\be
\sO'(x') := \sO'(\lam x) = \lam^{-\Delta_{\sO}} \sO(x) \label{deltadef} 
\ee
It is helpful to consider some examples:
\begin{itemize}[leftmargin=20pt, topsep=-1pt, itemsep=-1pt]
\item Consider a greyscale image viewed as a scalar field $p(x)$ where the value of $p(x)$ denotes the pixel value at point $x$. Under a conformal transformation only the argument of the field changes and not the value of the field; thus for image pixel data we would take $\Delta_{\mathrm{pixel}} = 0$.
\item Consider the electric field $\vec{E}$ in ordinary electrodynamics (which is a theory invariant under conformal transformations \cite{El-Showk:2011xbs}). The field in the vicinity of a point source obeys the inverse square law $\vec{E}(x) \sim \frac{\hat{x}}{|x|^2}$, which we see is consistent with \eqref{deltadef} if $\Delta_{E} = 2$, which is indeed the correct value. 
\end{itemize} 
In physical applications the values of $\Delta_{\sO}$ for a given conformally invariant system -- e.g. a particular kind of phase transition -- are often of great interest, as they are pure numbers that are {\it universal}: they generally do not depend on the microscopic details of the system, only on the symmetries and the spatial dimensionality. A quantitative understanding of the universality of these numbers was one of the drivers of the development of the {\it renormalization group}\footnote{See e.g. the 1982 Nobel Prize lecture by Wilson \cite{wilson1983renormalization} for a historical overview, or \cite{, cardy1996scaling, kardar2007statistical, zinn2021quantum} for textbook treatments.} one of the key insights of theoretical physics of the 20th century.

\begin{wrapfigure}{L}{0.4\linewidth}
    \centering
    \vspace*{-3em}
    \hspace*{3em}
    \input{figures/hyperboloid}
    \vspace*{-1em}
    \caption{Example of hyperboloid $ \|Y\| = -1$ embedded in $\mathbb{R}^{2,1}$, constituting two copies of $\AdS_2$.}
\end{wrapfigure}

\subsection{The Anti de Sitter space $\AdS_{d+1}$} 
To organize the data, we will instead lift it from $\mathbb{R}^{d}$ to an auxiliary space with {\it one higher dimension}, i.e. {\it Anti de Sitter space} $\AdS_{d+1}$. This space can conveniently be understood in terms of a submanifold of $\mathbb{R}^{d+1,1}$, equipped with its natural metric $\mathrm{diag}(-1,1,1\cdots,1)$. Consider the submanifold given by the constraint $\|Y\| = -1$, where $Y \in \mathbb{R}^{d+1,1}$. As shown in Figure \ref{fig:hyperboloid}, this submanifold has two connected components, each of which is defined to be a copy of $\AdS_{d+1}$.\footnote{For a more abstract and general definition of $\AdS$ for general quadratic spaces we refer to \Cref{def:dS-AdS} and the corresponding (partial) parameterization presented here to  \Cref{sec:AdS-param}.}
For concreteness, we will present all formulas working with the ``top'' branch, i.e. the one which has $Y^{0} > 0$, and we define:
\be
\AdS_{d+1} := \{ Y \in \mathbb{R}^{d+1,1} \;\;| \;\;\|Y\| = -1, Y^0 > 0\}. \label{hyp1} 
\ee

There is a natural action of $\POr(d+1,1)=\Or(d+1,1) / \{\pm 1\}$ on this submanifold, given by $\Or(d+1,1)$ rotations on $Y$ on each connected component. This forms the \emph{isometry group} $\Isom(\AdS_{d+1})$ of $\AdS_{d+1}$, 
see \cite{McK25} Theorem 7.5 or \ref{thm:isom-dS-AdS}. 
Note that $\POr(d+1,1)$ is also precisely the global conformal group $\GConf(\mathbb{R}^d)$ of $\mathbb{R}^d$. This is not an accident: the fact that local operations in the interior of $\AdS_{d+1}$ result in conformally invariant operations on $\mathbb{R}^d$ is very well known in the context of the AdS/CFT correspondence in quantum gravity \cite{Maldacena:1997re,Gubser:1998bc,Witten:1998qj}, which states that under some circumstances a quantum gravity theory on $\AdS_{d+1}$ is equivalent to a quantum field theory with conformal symmetry on $\mathbb{R}^d$. Also see \Cref{rem:isom-dS-AdS-conf-Rd}.

Here we will not use any of the dynamics of quantum gravity, but we will exploit some of the well-studied kinematics of that correspondence as a convenient tool to build convolutional kernels on $\AdS_{d+1}$. Indeed, a general framework for constructing convolutional layers that are equivariant under isometry groups of any pseudo-Riemannian manifold was given in \cite{weiler2023EquivariantAndCoordinateIndependentCNNs, zhdanov2024cliffordsteerable}, and our work can be viewed as building on a special case of that. 

To be more precise, we first place explicit coordinates $X=(X^1,\dots,X^{d+1})$ on $\AdS_{d+1}$. It will be convenient to separate these coordinates as $X=(X^1,\dots,X^{d+1}) = (x^1,\dots,x^d,z)=(x,z) \in \R^d \times \R_{>0}$ where $x \in \mathbb{R}^d$ and $z \in \R_{>0}$ is an extra dimension. We can now solve the hyperboloid constraint \eqref{hyp1} in terms of these coordinates as
\begin{equation}
Y^0  = + \frac{z}{2}\le(1 + \frac{1}{z^2}(1 + \sum_{a=1}^d x^a x^a)\ri)  > 0, \qquad
Y^a  = \frac{1}{z} x^{a}, \qquad Y^{d+1}  = \frac{z}{2}\le(1 - \frac{1}{z^2}(1 - \sum_{a=1}^d x^{a} x^{a})\ri).  
\label{ydp1} 
\end{equation} 
The Riemannian metric on $\AdS_{d+1}$ is the induced metric on this hyperboloid. A short computation, see \Cref{prp:para-ads}, gives:
\be
ds^2 = \sum_{\mu,\nu=1}^{d+1} g_{\mu\nu}(X) dX^{\mu} dX^{\nu} = \frac{1}{z^2}\le(\sum_{a=1}^{d} (dx^a)^2 + dz^2\ri). \label{adsmetric} 
\ee
Finally, we record how the isometry group acts on $\AdS_{d+1}$ as $X' = G X$. Using the same parameters as in \eqref{confgroup} we have: 
\be
(x',z') = (x + t, z) \qquad (x', z') = (M x, z) \qquad (x', z') = (\lambda x, \lambda z), \label{adsisom1} 
\ee
and
\be
\le(\frac{x'}{\|x'\|^2 + z'^2}, \frac{z'}{\|x'\|^2 + z'^2}\ri) = \le(\frac{x}{\|x\|^2 + z^2} + b, \frac{z}{\|x\|^2 + z^2}\ri). \label{adsisom2} 
\ee

A key point is that the manifold (\ref{adsmetric}) has a $d$-dimensional boundary at $z =0$. This boundary is mapped to itself under the isometries. Furthermore, the isometry group acts on the boundary points $(x^{a}, z = 0)$ precisely as in (\ref{confgroup}). Thus one should imagine that conformal data on $\mathbb{R}^d$ ``lives on the boundary of $\AdS_{d+1}$''. In what follows a key role will be played by the $\POr(d+1,1)$-invariant proper distance between $D(X,X')$ between two points in $\AdS_{d+1}$, which is related to the absolute inner product $|\langle Y, Y'\rangle|$ and is given explicitly by:
\be
\label{eq:proper_distance}
\cosh D(X, X') := |\langle Y, Y'\rangle| \overset{!}{=} \frac{z^2 + z'^2 + \sum_{a=1}^{d}(x^a - x'^a)^2}{2 z z'}.
\ee

\section{AdS-GNN} 

\begin{wrapfigure}{R}{0.4\linewidth}
    \vspace{-5em}
    \centering
    \input{figures/lifting}
    \vspace{-3.5em}
    \caption{An example of the embedding of a boundary digit into $\AdS$.}
    \label{fig:embedding-example}
    \vspace{-5em}
\end{wrapfigure}

We will now describe how to use these ideas to formulate a conformally equivariant neural network by extending the data from the boundary into the bulk of $\AdS_{d+1}$. It is possible to do this formally for conformal fields $\phi(x^{a})$ living on $\mathbb{R}^{d}$, resulting in a field $\Phi(x^{a};z)$ living on $\AdS_{d+1}$. However our preliminary experiments show that discretization issues and boundary effects make it difficult to control equivariance errors in that case. In this work we will study instead point clouds on $\mathbb{R}^d$.  

\subsection{Embedding points in $\AdS$}
Consider a point cloud of $N$ points in $\mathbb{R}^{d}$, $\{ x_i \}$, $i \in \{1, \cdots, N\}$. We would now like to lift this data into the bulk of $\AdS_{d+1}$ in a manner that preserves the symmetries. 

A first attempt from the correspondence of symmetries shown in (\ref{adsisom1}) is to simply embed each of the points directly into the boundary $z = 0$, i.e. with $X^{\mu}_i = (x^{a}_i, z = 0)$. 
However the metric (\ref{adsmetric}) has a singularity at $z = 0$ -- e.g. from (\ref{eq:proper_distance}) note that each point at $z = 0$ is at infinite proper $\AdS$ distance from any points at $z > 0$ -- and thus such an attempt will require us to pick a regulating value of $z$. Using a fixed constant will explicitly break the symmetries.

We instead determine a value of $z_i$ for each point using an approach outlined in Algorithm \ref{alg:ads_embedding}. We first embed each point into $\AdS$ using
\be
X^{\mu}_i = (x^{a}_i, z = z_0)
\ee
with a small regulator $z_0$. For each point, we then compute the {\it AdS center of mass} $\hat{X} = (\hat{x}_i, \hat{z}_i)$ of its $k_{\mathrm{lift}}$ nearest neighbours. The $\AdS$ center of mass is a generalization to hyperbolic space of the familiar notion of the center of mass from flat Euclidean space. It can be computed easily using an approach due to Galperin \cite{galperin1993concept} which we review in Appendix \ref{app:galperin}.

The geometry of $\AdS$ implies that the center of mass will generally be deeper inside than the original points. Importantly the $z$ value of the centroid now has a finite limit as $z_0 \to 0$, in which case it depends on the (appropriately averaged) relative separation of the points. We then perform a final embedding of the point using this $z$ value, i.e.
\be
X_i^{\mu} = (x_i^{a}, \hat{z}_i) \ . \label{embval} 
\ee

\begin{wrapfigure}{R}{0.53\linewidth}
\vspace{-23pt}
\begin{minipage}{\linewidth}
\begin{algorithm}[H]
\caption{AdS Embedding}\label{alg:ads_embedding}
\centering
\begin{algorithmic}[1]
\Require $X = \{\mathbf{x}_i\}_{i=1}^N \subset \mathbb{R}^d$, $k_\mathrm{lift} \in \mathbb{N}$, $z_0 \in \mathbb{R}$
\For{each point $i \in \{1,\ldots,N\}$}
\State $z_i \gets z_0$
\State $\text{neighbors}_i \gets \text{KNN}(x_i, X, k_\mathrm{lift})$
\State $(\hat{x}_i, \hat{z}_i) \gets \text{ComputeAdSCoM}(\text{neighbors}_i)$
\EndFor
\\
\Return $\{(x_i, \hat{z}_i)\}_{i=1}^N \subset AdS_{d+1}$
\end{algorithmic}
\end{algorithm}
\end{minipage}
\vspace{-5pt}
\end{wrapfigure}

Intuitively, the $z$ coordinate corresponds to the {\it length scale} of the degrees of freedom we are considering\footnote{This is familiar from the physics of the AdS/CFT correspondence, where it is well-understood that the infrared physics lives deeper in the bulk \cite{Susskind:1998dq}.}. Our choice above amounts to saying that the appropriate length scale for a point $x_i$ is related to its distance from its neighbours. This exactly preserves scale invariance, but it gently breaks special conformal transformations. This is expected on physical grounds, as generally any choice of regulator necessarily breaks conformal invariance \cite{cardy1996scaling}. In our experiments, we check generalization under special conformal transformations empirically and verify that the breaking is mild.   

Finally, we have discussed lifting the points $x_i$. The input data may also have some {\it features} $h^{\mathrm{input}}_i$ associated to each point $x_i$. They should be interpreted as a sample of an underlying conformal field $\sO(x)$ with dimension $\Delta$, where $h^{\mathrm{input}}_i = \sO(x_i)$. This boundary feature should be contrasted with the feature associated with a bulk point in $\AdS$; this is a scalar and does not transform with an associated factor of $\lam^{\Delta}$ as in \eqref{deltadef}. Said differently the full dependence of bulk features under scaling arises from its dependence on an extra coordinate $z$. This difference in representation is taken into account by lifting the feature as follows
\be
h^{\mathrm{lifted}}_i = \hat{z}_i^{\Delta} h^{\mathrm{input}}_i \label{liftfeat} 
\ee
For many cases (e.g. image data) the input feature will have $\Delta = 0$ and this step may be skipped. This relation has an analogue in terms of bulk-to-boundary propagators from the AdS/CFT correspondence, where such factors of $z^{\Delta}$ relate physics in the bulk (i.e. $h_{i}^{\mathrm{lifted}}$) to that of the boundary (i.e. $h_{i}^{\mathrm{input}}$)\footnote{See e.g. Section 2.5 of \cite{Witten:1998qj} for more details.} 

\subsection{Message passing}
\vspace{-5pt}
Given this set of points $\{X_i\}$ in $\AdS$, we now operate on it using a graph neural network. 

To orient ourselves, we recall first an earlier model, that of E(n) Equivariant Graph Neural Networks (EGNNs) \cite{SatorrasHW21, liu2024clifford}. These are graph neural networks that are equivariant to flat-space rotations, translations, reflections and permutations. The input to the model is a graph $G = (\mathcal{V},\mathcal{E})$ whose vertices are embedded into Euclidean space $\mathbb{R}^d$. We denote the position of node $v_i$ as ${\bf p}_i \in \mathbb{R}^d$ and its latent $D$-dimensional feature vector of node $v_i$ as $\mathbf{h}_i$.  
The $l$-th layer of EGNN is then defined as
\begin{align}
\label{eq:egnnlayer}
    &\mathbf{m}_{ij} = \psi_e(\mathbf{h}_i^l, \mathbf{h}_j^l, \|\mathbf{p}_i - \mathbf{p}_j\|_2), \qquad &\textrm{EGNN message} \\
    &\mathbf{h}_i^{l+1} = \psi_h(\mathbf{h}_i^l, \mathbf{m}_i), \quad \mathbf{m}_i = \sum_{j\in \mathcal{N}(i)} \mathbf{m}_{ij}, \nonumber \qquad &\mathrm{aggregate + update}
\end{align}
where here $\mathcal{N}(i)$ represents the set of neighbours of node $v_i$, $\psi_e, \psi_h$ are message and update MLPs which we see are conditioned only on the pairwise distance in $\mathbb{R}^d$ betwen nodes. 

We thus define {\bf AdS-GNN}: we adopt the model above to operate on $\AdS$ where a graph $G$ is embedded. As above, each node $v_i$ has a position $X_i \in \AdS_{d+1}$ and a latent feature vector ${\bf h_i}$. If edges $\mathcal{E}$ are not provided in the graph, we induce connectivity with $k_{\text{con}}$ nearest neighbours using the $\AdS$ proper distance~(\ref{eq:proper_distance}).

In the message function~(\ref{eq:egnnlayer}), we also use the $\AdS$ proper distance instead of the Euclidean one, i.e.
\begin{align}
\label{eq:ads_message}
    &\mathbf{m}_{ij} = \psi_e(\mathbf{h}_i^l, \mathbf{h}_j^l, D(X_i,X_j)), \qquad &\textrm{AdS-GNN message}
\end{align}
which yields an efficient conformal group equivariant GNN without substantial computational overhead compared to its Euclidean counterpart. Note that by conditioning on AdS proper distance, we introduce a notion of locality both in ordinary space and in scale (as represented by the $z$ coordinate). Though the embedding of the point cloud mildly breaks special conformal transformations, the graph neural network itself is exactly invariant under all of $\Isom(\AdS_{d+1})$.

The framework above will result in features ${\bf h_i}$ that are manifestly invariant under the conformal group. If performing an invariant task (e.g. classification) we can now aggregate the information in a permutation-invariant manner by summing over nodes as usual at the final layer. On the other hand, for a regression task we should specify the transformation properties of the output; e.g. if the output from the network is a conformal field $\sO(x_i)$ living on the boundary it must transform under scale transformations with a specified $\Delta$ as in \eqref{deltadef}. This can be accomplished by taking the output $\sO(x_i)$ to be
\be
\sO(x_i) = \hat{z}_{i}^{-\Delta} h^{l_{\mathrm{final}}}_{i} \label{nontrivdef} 
\ee
where $h^{l_{\mathrm{final}}}_{i}$ is the conformally invariant output from the node associated with $x_i$ and $\hat{z}_i$ the $z$-coordinate of the embedding found in \eqref{embval}. This relation is the inverse tranformation of \eqref{liftfeat}, and guarantees that the output satisfies \eqref{deltadef}.

\begin{figure*}[t]
\vspace{-35pt}
\begin{minipage}[t]{0.6\textwidth}
\begin{table}[H]
\caption{Classification error on SuperPixel MNIST. We augment the test set with random rotations and scaling. \ding{55} indicates random-guess performance. Baseline results are taken from \cite{BekkersVHLR24}.}
\label{table:mnist}
\begin{center}
\scalebox{0.9}{
\begin{tabular}{lccc}
\hline
\multirow{2}{*}{\bf Model} & \multicolumn{3}{c}{\bf Error rate, \%} \\
& non-augmented & rotated & rotated+scaled \\ \hline
MONET & $8.89$ & \ding{55} & \ding{55} \\
SplineCNN & $4.78$ & \ding{55} & \ding{55} \\
GCGP & $4.2$ & \ding{55} & \ding{55} \\
GAT & $3.81$ & \ding{55} & \ding{55} \\
PNCNN & $1.24 \pm 0.12$ & \ding{55} & \ding{55} \\
P$\Theta$NITA & $1.17 \pm 0.11$ & $1.17$ & \ding{55} \\
EGNN & $4.17 \pm 0.45$ & $4.17$ & \ding{55} \\
AdS-GNN (\textbf{Ours}) & $4.09 \pm 0.27$ & $4.09$ & $4.09$ \\
\hline
\end{tabular}}
\end{center}
\end{table}
\hfill
\vspace{-5pt}
\end{minipage}
\hfill
\begin{minipage}[t]{0.3\textwidth}
\begin{figure}[H]
\caption{Test error on augmented data, SuperPixel MNIST.}
\label{fig:eq-error}
\begin{center}
\includegraphics[width=\linewidth]{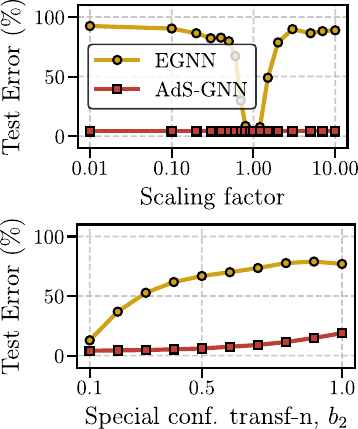}
\end{center}
\end{figure}
\end{minipage}
\vspace{-20pt}
\end{figure*}

\section{Experimental results}
We test the framework above on tasks that are loosely divided into two types: computer vision tasks and an application to the Ising model in statistical physics. 
\vspace{-5pt}

\subsection{Computer vision tasks}

\paragraph{SuperPixel MNIST} 
We benchmark AdS-GNN on the super-pixel MNIST dataset \cite{MontiBMRSB17}, which consists of 2D point clouds of MNIST digits segmented into $75$ superpixels. Results are given in Table~\ref{table:mnist}. For in-distribution data, AdS-GNN performs on par with its roto-equivariant counterparts. It does however fall slightly behind P$\Theta$NITA, which has orientational information; we feel this happens as AdS-GNN is unable to handle orientation and relies on invariant descriptors. We also study the response of a model to various augmentations (see Fig.~\ref{fig:eq-error}); for this we compare to EGNN and find that AdS-GNN has much stronger generalization capacities. As expected, AdS-GNN is precisely scale-invariant. For special conformal transformations, there is a small breaking of symmetry arising from the uplift. We empirically measure the effect on augmented performance in Figure \ref{fig:eq-error} for a special conformal transformation as in (\ref{confgroup}) parametrized by $b = (0, b_2)$ and verify that it is very small.   %

\vspace{-10pt}

\paragraph{Shape segmentation}
We further benchmark against a shape recognition dataset, where we sample points from a selection of randomly sized and rotated shapes (either a square, a circle, or a triangle). The task is to assign a shape class to each point in the cloud. We benchmark against EGNN and a message-passing neural network MPNN which conditions messages against $x_i - x_j$ (and thus has only translational equivariance, not rotational). Here, as shown in the left panel of Figure \ref{fig:isingloss} we find that AdS-GNN outperforms EGNN even on in-distribution performance. We believe that this happens as the training data contains structure at different sizes to which AdS-GNN adapts more efficiently than EGNN, particular when there is a small number of training points. Both of them fall behind MPNN, which we again attribute to their lack of orientational information. 

A further segmentation task -- PascalVOC -- is discussed in the Appendix, where we find that AdS-GNN is essentially the same as EGNN.

\subsection{Ising correlation functions} We next consider a task from statistical physics, that of predicting $N$-point correlation functions in the 2d Ising model. As we review in more detail in Appendix \ref{app:ising-review}, this is a simple model of magnetism with a tunable temperature; as one increases it the model undergoes a phase transition from a ferromagnetic to a paramagnetic phase. At the phase transition point, fluctuations of the magnetization exhibit conformal symmetry and are precisely understood in terms of a continuum model called the 2d Ising conformal field theory. This model turns out to have two non-trivial types of conformal field, the spin $\sig(x)$ and the energy $\ep(x)$, which have conformal dimensions $\Delta_\sig = \frac{1}{8}$ and $\Delta_{\ep} = 1$ respectively\footnote{There are other slightly non-local fields that we will not discuss here.}. This model is exactly solvable, and the $N$-th moment of the energy operator\footnote{This relation holds for even $N$; for odd $N$ the correlation function vanishes.} can be computed to be
\be
\vev{\ep(\zeta_1) \ep(\zeta_2) \cdots \ep(\zeta_N)} = \bigg|\mathrm{Pf}\le[\frac{1}{\zeta_i - \zeta_j}\ri]_{1 < i,j < N}\bigg|^2 \label{pfaffian} 
\ee
with $\mathrm{Pf}$ the Pfaffian and where $\zeta_i = x_i + i y_i$ is a representation of points in $\mathbb{R}^{2}$ as complex numbers. (As we describe in Appendix \ref{app:ising-review}, there is a corresponding formula (\ref{spincorr-app}) for the moment $\vev{\prod_{i=1}^{N} \sig(\zeta_i)}$ of the spin operator). The compact form of the answer hides the fact that this is a very complicated expression. These functions have an erratic pattern of spikes when two points come close, on top of a more gentle modulation arising from the background of the other points. 

The task is to predict the (logarithm of) the $N$-th moment of the energy and spin operators as a function of the input points. We create a training dataset where we sample the coordinates of input points uniformly in $[-2, 2]$ for various values of $N$ and use the correlation functions from \eqref{pfaffian} as the training data. We simultaneously predict the spin and energy moments, using the sum of relative L2 losses. As the output is the $N$-point function of conformal fields with a non-trivial $\Delta$ we use $N$ copies of \eqref{nontrivdef} to find our prediction $\mathrm{Pred}_{a}(\{x_i\})$ in each channel $a$ to be
\be
\log(\mathrm{Pred}_{a}(\{x_i\})) = \mathrm{AdSGNN}_{a}(\{x_i\})-\Delta_{a} \sum_{i=1}^{N} \log(\hat{z}_i)  \label{modDelta} 
\ee
where $a \in \{\sig, \ep\}$ and the $\Delta_{a}$'s are trainable parameters which can be interpreted as the learned conformal dimension of the $\sig$ and $\ep$ fields respectively.  

\begin{figure}
  \centering
  \begin{subfigure}[b]{0.3\textwidth}
    \centering
    \includegraphics[width=\textwidth]{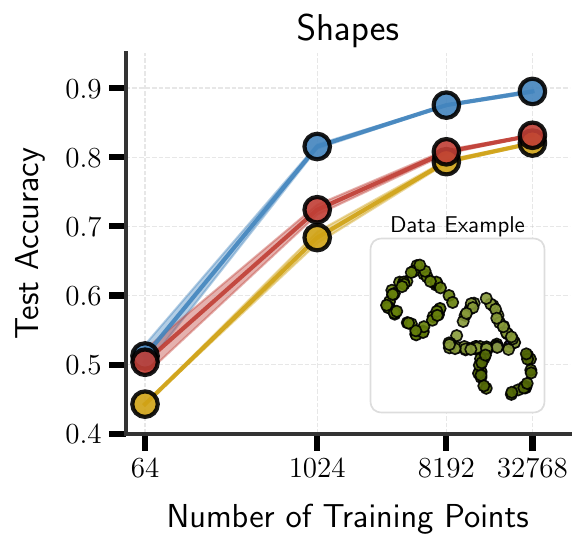}
  \end{subfigure}
  \hfill
  \begin{subfigure}[b]{0.3\textwidth}
    \centering
    \includegraphics[width=\textwidth]{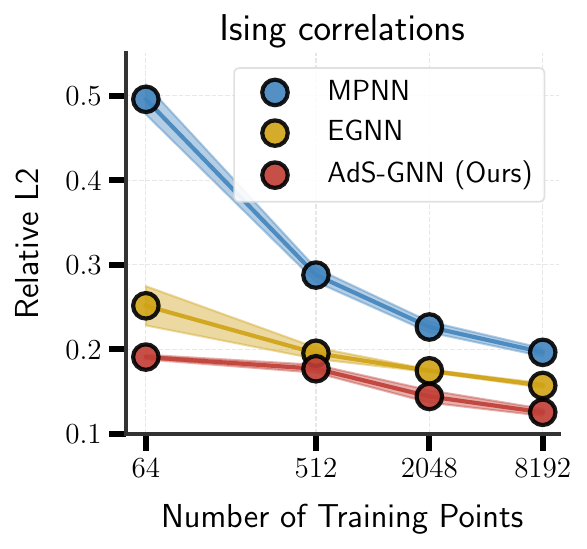}
  \end{subfigure}
  \hfill
  \begin{subfigure}[b]{0.3\textwidth}
    \centering
    \includegraphics[width=\textwidth]{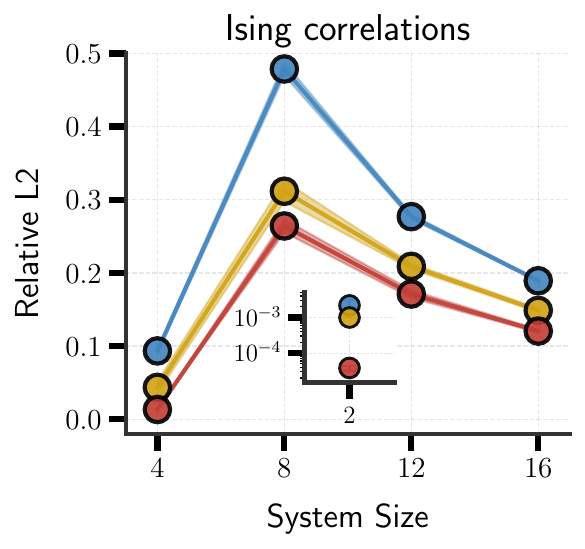}
  \end{subfigure}
  \caption{Performance on shape segmentation and the Ising task. {\it Left}, test accuracy of shape segmentation as a function of the number of training points. {\it Middle}, relative L2 as a function of the number of training points, with system size fixed at $N = 16$. {\it Right}, loss as a function of system size with 8192 training points. Inset shows $N = 2$.}
  \label{fig:isingloss}
  \vspace{-1.5em}
\end{figure}

We benchmark against EGNN and a baseline message-passing neural network (MPNN) whose messages are conditioned against $x_i - x_j$, and which thus only has translational equivariance. The results are shown in Figure \ref{fig:isingloss} for the choice $k_{\mathrm{lift}} = 1$. We see that the performance of AdS-GNN is {\it superior in all regimes}. We also note various useful features.

\begin{wraptable}{r}{0.55\textwidth}
\vspace{-1em}
\begin{tabular}{c c c}
\textbf{System  size} & \textbf{$\Delta_{\ep}$ (energy)} & \textbf{$\Delta_{\sig}$ (spin)}  \\
\hline
2 & $1.0000 \pm 0.0000$ & $0.1250 \pm 0.000$ \\
4 & $0.9998 \pm 0.0000$ & $0.1250 \pm 0.000$ \\
8 & $0.9924 \pm 0.0010$ & $0.1248 \pm 0.000$ \\
12 & $0.9894 \pm 0.0032$ & $0.1247 \pm 0.000$ \\
16 & $0.9893 \pm 0.0007$ & $0.1247 \pm 0.000$
\end{tabular}
\caption{Learned value of $\Delta$'s from AdS-GNN. Ground truth values are $\Delta_{\ep} = 1$ and $\Delta_{\sig} = 0.125$. Statistical uncertainties are standard deviations from 5 runs; for $\Delta_{\sig}$ they are $O(10^{-5})$ and are not quoted.}
\label{tab:delta} 
\vspace{-3em}
\end{wraptable}

\paragraph{Interpretability:} from \eqref{modDelta}, the learned values of the conformal dimensions $\Delta_{\ep}$ and $\Delta_{\sig}$ may be read off at the end of training, as shown in Table \ref{tab:delta}. They are very close to the ground truth, showing the useful ability of this model to extract conformal dimensions from data. 

\paragraph{$N = 2$:} Note that for $2$ points (inset on right panel of Figure \ref{fig:isingloss}) AdSGNN performs more than an order of magnitude better than the others. This is because the form of the two-point function is completely fixed by conformal invariance to be $|x_1 - x_2|^{-2\Delta}$, and thus AdS-GNN need only learn a single number, whereas the others must learn to reconstruct the functional form of the power law.

\begin{figure}
\centering
\begin{subfigure}[b]{0.3\textwidth}
    \centering
    \includegraphics[width=\linewidth]{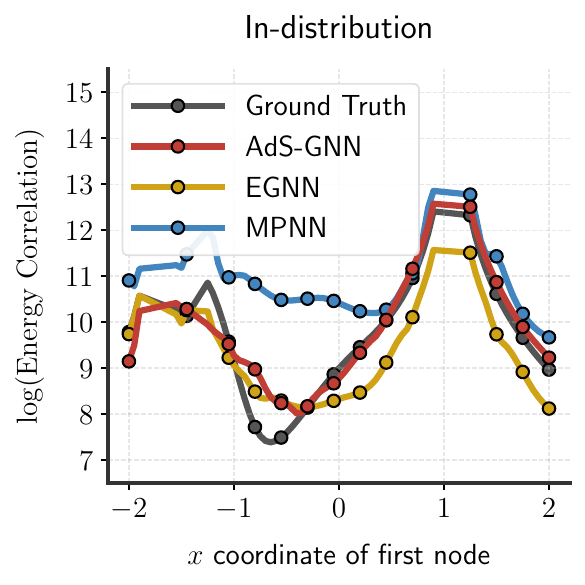}
\end{subfigure}
\begin{subfigure}[b]{0.3\textwidth}
    \centering
    \includegraphics[width=\linewidth]{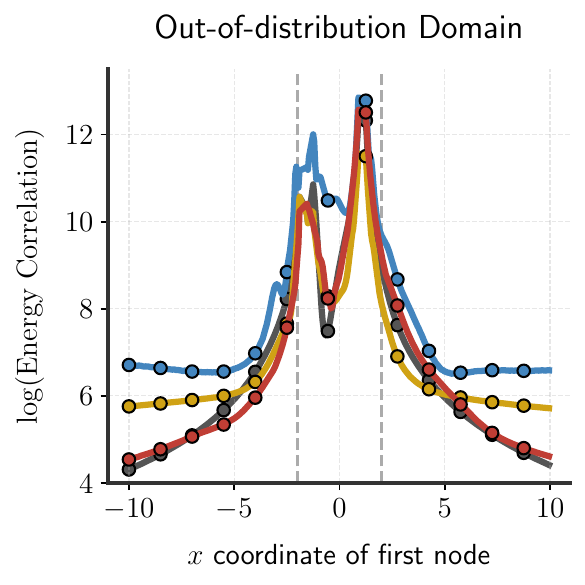} 
    \end{subfigure}
\begin{subfigure}[b]{0.3\textwidth}
    \centering
    \includegraphics[width=\linewidth]{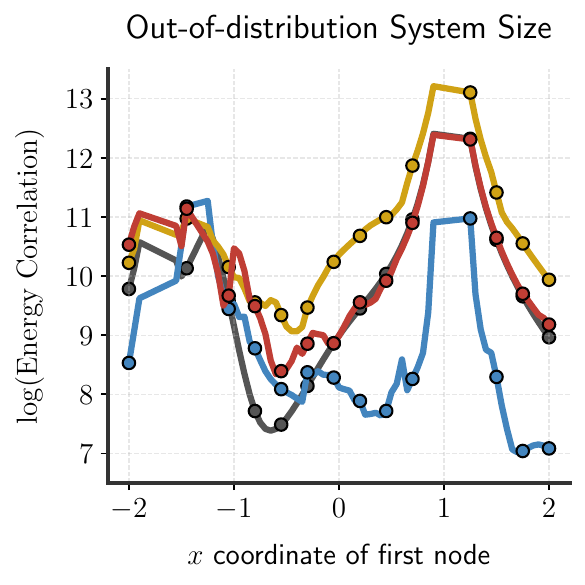}
\end{subfigure}
\caption{Visualization of the output from various models on the Ising task; here all points are fixed except for the $x$ coordinate of the first, which is varied. {\it Left}, models tested on in-distribution data. {\it Center}, testing on values of $x$ which are far out of the training range, which is shown with dashed lines. {\it Right}, testing a model trained on $N = 8$ on a test set with $N = 16$.} 
\label{fig:encorrwithx}
\end{figure}

\paragraph{Generalization:} AdS-GNN generalizes better than EGNN in two different directions. As shown in Figure \ref{fig:encorrwithx}, it generalizes well to values of $x$ that are outside the training data, as one might have expected from scale-equivariance. AdS-GNN also generalizes well {\it across $N$}; as shown in an example in Figure \ref{fig:encorrwithx}, AdS-GNN trained with one value of $N$ works well when asked to predict correlation functions with a different $N$, suggesting that it has robustly understood the underlying physics. It always generalizes better than EGNN, particularly when the graph connectivity is more dense. This is discussed quantitatively in Appendix \ref{app:ising-gen}.

\section{Conclusion}
\vspace{-5pt}
In this paper, we introduced AdS-GNN - a neural network that is equivariant with respect to conformal transformations. We have demonstrated strong performance on various tasks. We found particularly interesting an application to the Ising model, where the model exhibited impressive generalization capacities and {\it interpretability}, in that the conformal dimensions $\Delta_{a}$ -- important universal quantities -- could be extracted from the trained model. One might ask what other such universal information exists: a general conformally invariant field theory in physics is defined in terms of these dimensions $\Delta_{a}$ and a set of 3-point coefficients $c_{abc}$ which turn out to determine higher-order moments such as \eqref{pfaffian}\footnote{See e.g. \cite{DiFrancesco:1997nk,Simmons-Duffin:2016gjk} for textbook treatments.}. We expect that the information on the $c_{abc}$ is encoded in the successfully predictive models trained above, and it is very interesting to ask how it could be usefully extracted. We anticipate further applications both to critical phenomena and long-range interactions in physical systems and to tasks in computer vision that require robust characterization of shape and form.

\subsection{Limitations}
One limitation of our work is that while the operations after lifting to $\AdS_{d+1}$ are equivariant under the full conformal group, the lifting procedure from $\mathbb{R}^d$ to $\AdS_{d+1}$ itself is equivariant only under rotations, translations, and scaling, and mildly breaks special conformal transformations. Another limitation is that we only use scalar features in the construction.

\begin{ack}
MZ was supported by
Microsoft Research AI4Science. This work was supported by a grant from the Simons Foundation (PD-Pivot Fellow-00004147, NI and EB). NI is supported in part by the STFC under grant number ST/T000708/1. This publication is part of the project SIGN with file number VI.Vidi.233.220 of the research programme Vidi which is (partly) financed by the Dutch Research Council (NWO) under the grant https://doi.org/10.61686/PKQGZ71565
\end{ack}

\newpage
\bibliography{refs}
\bibliographystyle{utphys}

\newpage

\appendix
\section{$\AdS_{d+1}$ Details}
We provide a few further details about aspects of $\AdS_{d+1}$: 

\subsection{Action of group}
Recall that $\AdS_{d+1}$ can be understood as the hyperboloid defined in \eqref{hyp1}. The action of $[\Lambda] \in \Or(d+1,1)/\{\pm I \}$ then works concretely as follows:
$$ [\Lambda].Y := \mathrm{sign}((\Lambda Y)^0) \cdot (\Lambda Y) ,  $$
where $\Lambda Y$ is just vector matrix multiplication and where the scalar multiplication with $\mathrm{sign}((\Lambda Y)^0)$ corrects the sign of the $0$-th entry (the component of $(-1)$-signature). 
Together we get:
\be
(([\Lambda].Y)^0)^2 - \sum_{i=1}^{d+1} (([\Lambda].Y)^i)^2 = 1, \qquad ([\Lambda].Y)^0 > 0,
\ee
and thus: $[\Lambda].Y \in \AdS_{d+1}$. 
This gives us a well-defined group action of $\Or(d+1,1)/\{\pm I \}$ on $\AdS_{d+1}$, and, in particular, a well-defined group action of $\Conf(\mathbb{R}^d)$ on $\AdS_{d+1}$.

\subsection{The center of mass of a set of points on AdS} \label{app:galperin}
We will require an expression for the ``center of mass'' $C(\{X_i\})$ of a set of points on AdS. This problem was solved in \cite{galperin1993concept}; the basic idea is to view the hyperboloid as a submanifold of $\mathbb{R}^{d+1,1}$ as above, use additivity properties there to find a vector, and then find the intersection of the ray in the direction of that vector with the hyperboloid. 

In practice, this is quite simple to implement. Denote the center of mass by $\bar{Y}^{A}$, and the set of $N$ points for which we want the centroid by $(x_{i}^{a}, z_{i})$. We would like to find the analogous coordinates for the centroid $(\bar{x}^a, \bar{z})$. 

We have that
\begin{align} 
\bar{Y}^0 & \equiv  \frac{\bar{z}}{2}\le(1 + \frac{1}{\bar{z}^2}(1 + \sum_{a} \bar{x}^a \bar{x}^a)\ri) = \frac{1}{\mathcal{N} N} \sum_{i} \frac{z_i}{2}\le(1 + \frac{1}{z_i^2}(1 + \sum_{a} x_i^a x_i^a)\ri)\\
\bar{Y}^{a} & \equiv \frac{\bar{x}^{a}}{\bar{z}} = \frac{1}{\mathcal{N} N} \sum_{i} \frac{x_i^{a}}{z_i} \\ \bar{Y}^{d+1} & \equiv \frac{\bar{z}}{2}\le(1 - \frac{1}{\bar{z}^2}(1 - \sum_{a} \bar{x}^{a} \bar{x}^{a})\ri) = \frac{1}{\mathcal{N} N} \sum_{i}\frac{z_i}{2} \le(1 - \frac{1}{z_i^2}(1 -\sum_{a} x_i^{a} x_i^{a})\ri)
\end{align}
The first equality is the definition of the embedding, the second is the definition of the centroid from Galperin. Here $\mathcal{N}$ is a normalization constant which is picked to guarantee that 
\be
(\bar{Y}^0)^2 - \sum_{a} (\bar{Y}^a)^2 - (\bar{Y}^{d+1})^2 = 1
\ee

So to find the centroid, the easiest thing to do is to compute the sums on the right hand side of the second equality, which thus determines the vector $\bar{Y}^{A}$ up to an overall scale  $\mathcal{N}$; then we enforce the norm constraint above which lets us find $\mathcal{N}$ and thus fixes the vector $\bar{Y}^{A}$ completely. We then express the answer in useful coordinates by solving for $(\bar{z}, \bar{x}^a)$ through %
\be
\bar{z} = \frac{1}{\bar{Y}^0 - \bar{Y}^{d+1}}, \qquad 
\bar{x}^a  = \frac{\bar{Y}^a}{\bar{Y}^0 - \bar{Y}^{d+1}} \ . 
\ee

To get some intuition for the procedure, we study it in the case of two points $X_1 = (x_1^{a}, \epsilon)$ and $X_2 = (x_2^{a}, \epsilon)$ starting at the same value of the $z$ coordinate. We find
\be
C(X_1, X_2) = \le(\ha(x_1^a + x_2^a), \ha\sqrt{|x_1 - x_2|^2 + 4\ep^2}\ri) \label{comexample} 
\ee
i.e. we simply take the average of the spatial coordinates and move inwards in $z$ by an amount which depends on the separation between the two points in the spatial direction. In this case the center of mass is actually the midpoint of the geodesic that connects the two points.

\section{Further experimental details}

 \subsection{PascalVOC-SP} In a further experiment, we also compare AdS-GNN to EGNN on the LRGB data \cite{DwivediRGPWLB22}, see Table~\ref{table:pascal}. This is a pixel segmentation task; thus the output data at each node is conformally {\it in}variant and we take $\Delta = 0$ in \eqref{nontrivdef}, i.e. we read out the output from ${\bf h}_i$ directly. The difference in performance between EGNN and AdS-GNN is statistically insignificant, which indicates that in this task conformal equivariance does not constrain the model significantly and still allows for high expressivity.

 \begin{table}[H]
\caption{Classification error on PascalVOC-SP.}
\label{table:pascal}
\begin{center}
\begin{tabular}{lcc}
\hline
\multicolumn{1}{c}{Model}  &\multicolumn{1}{c}{EGNN} &\multicolumn{1}{c}{AdS-GNN} \\
Test F1 $\mathbf{\uparrow}$ & $27.80 \pm 0.74$ & $28.07 \pm 0.57$ \\
\hline
\end{tabular}
\end{center}
\end{table}

 \subsection{Ising generalizability} \label{app:ising-gen}
 
 Here we discuss further the generalization properties of the Ising task discussed in the main text. In particular, we provide information on how a a model trained on a given number of nodes $N_{\mathrm{train}}$ performs when evaluated on a different number of nodes $N_{\mathrm{test}}$. The connectivity on the graph on $\AdS$ is controlled by a parameter $k_{\mathrm{con}}$, the number of nearest neighbours which are connected to each node. 
 
\begin{figure}[htp]
  \centering
  \begin{subfigure}[b]{0.45\linewidth}
    \centering
    \includegraphics[width=\linewidth]{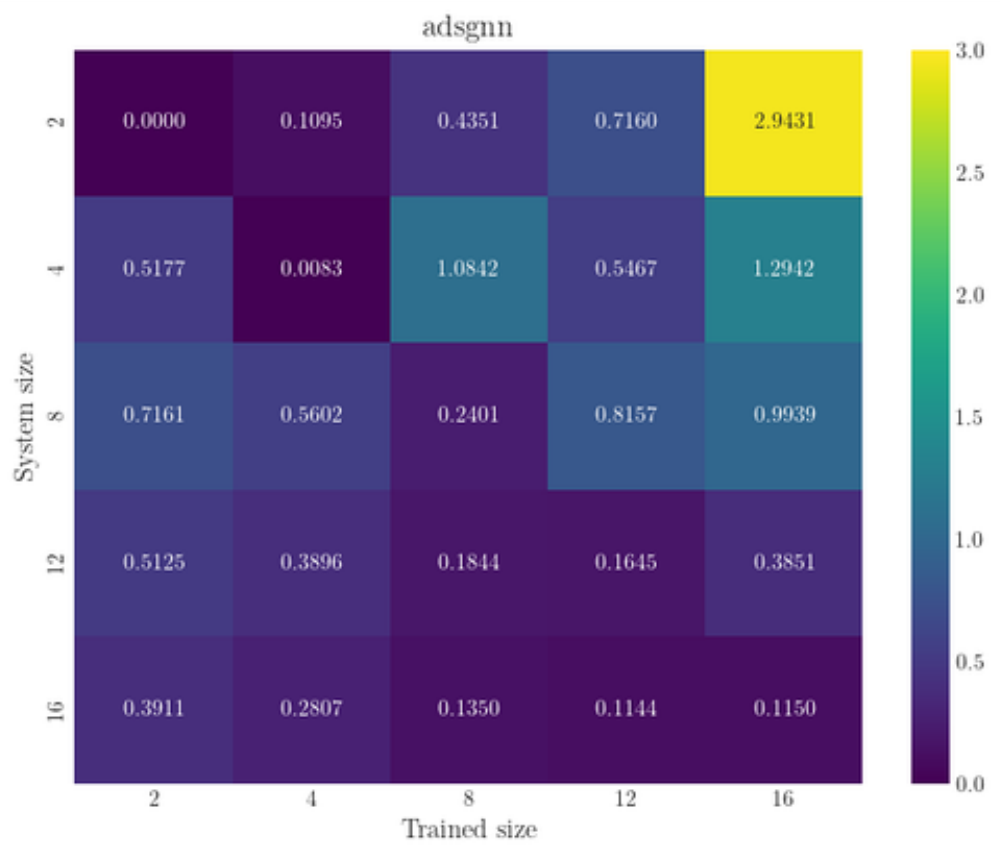}
    \caption{AdS-GNN with full connectivity}
    \label{fig:adsgnnfull}
  \end{subfigure}
  \hfill
  \begin{subfigure}[b]{0.45\linewidth}
    \centering
    \includegraphics[width=\linewidth]{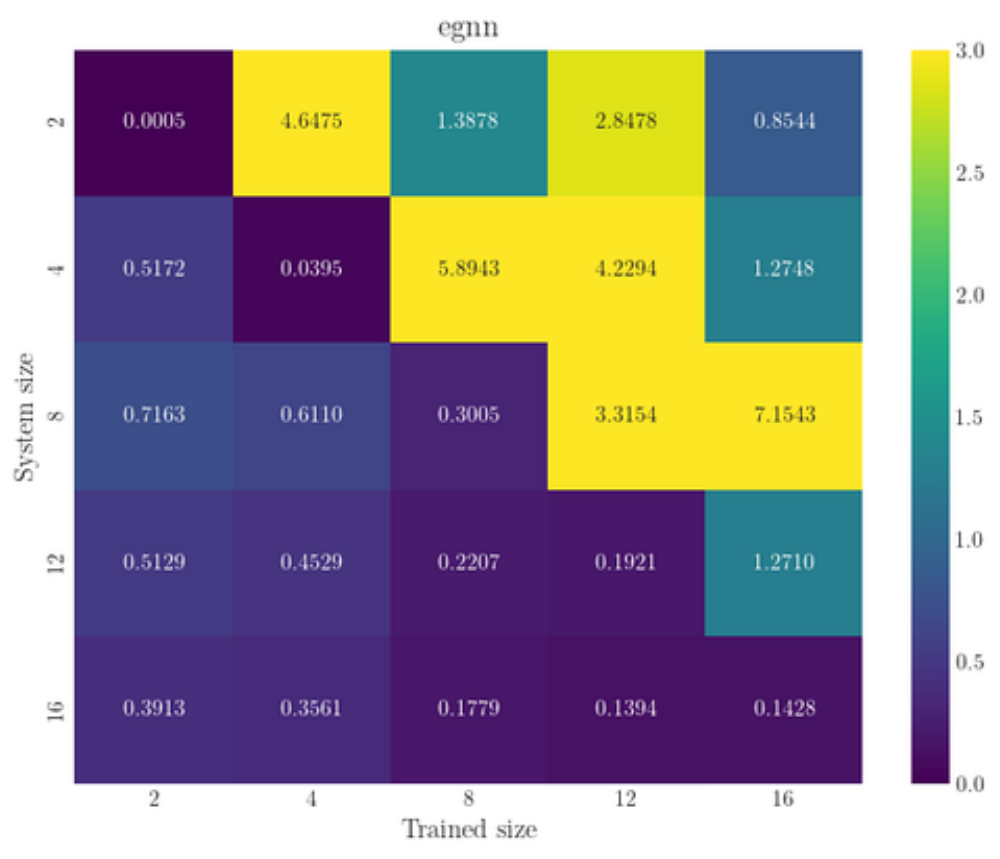}
    \caption{EGNN with full connectivity}
    \label{fig:egnnfull}
  \end{subfigure}

  \vspace{1em}  %

  \begin{subfigure}[b]{0.45\linewidth}
    \centering
    \includegraphics[width=\linewidth]{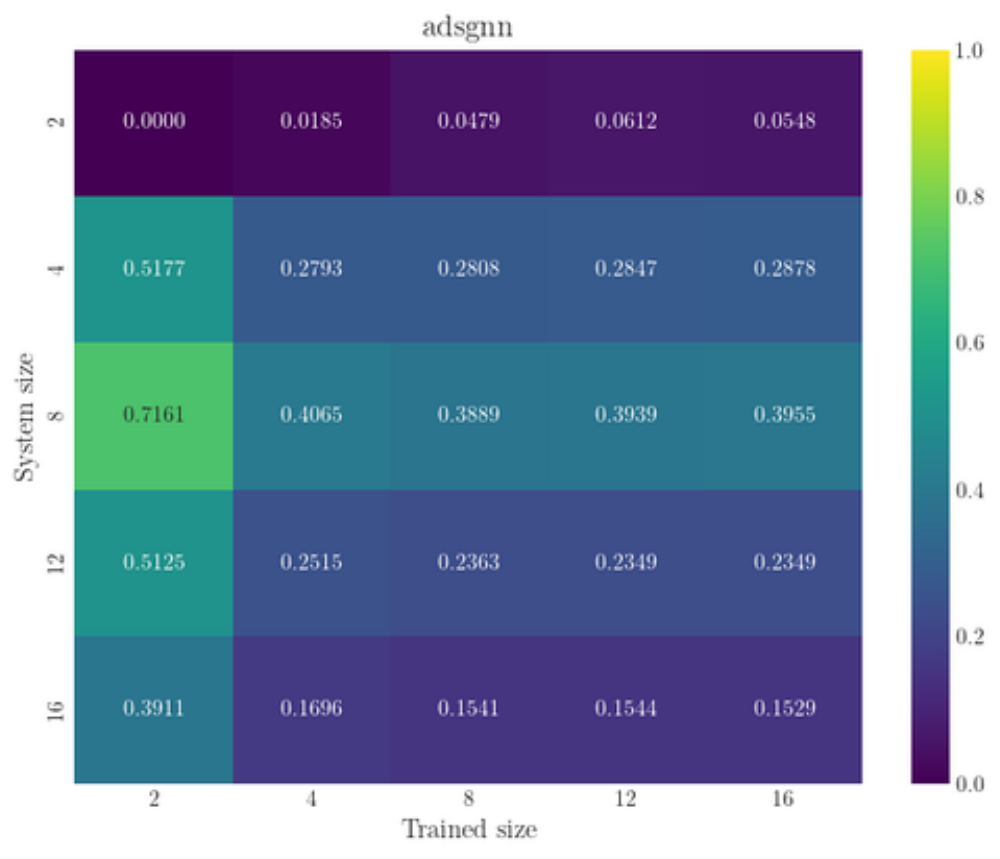}
    \caption{AdS-GNN with $k_{\mathrm{con}} = 2$}
    \label{fig:adsgnn-k-2}
  \end{subfigure}
  \hfill
  \begin{subfigure}[b]{0.45\linewidth}
    \centering
    \includegraphics[width=\linewidth]{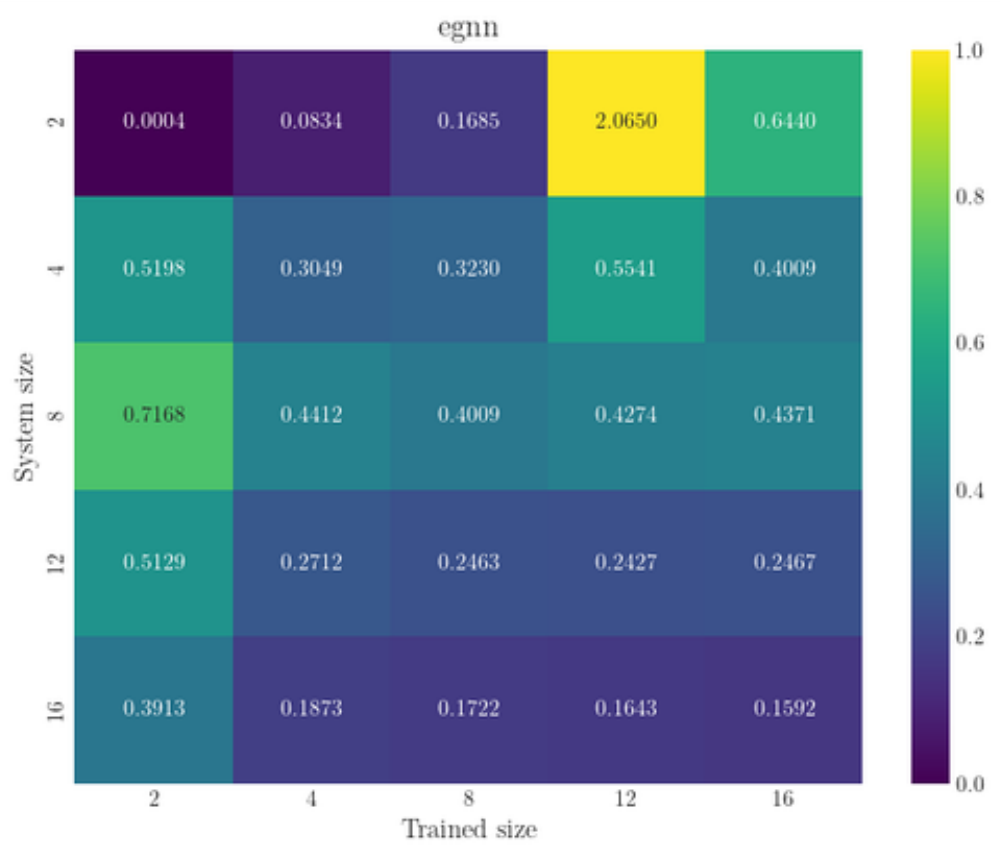}
    \caption{EGNN with $k_{\mathrm{con}} = 2$}
    \label{fig:egnn-k-2}
  \end{subfigure}

  \caption{Generalization across system size; each square shows the relative L2 loss of a system trained on a system of size $N_{\mathrm{train}}$ ($x$-axis) when tested on a system of size $N_{\mathrm{test}}$ ($y$-axis) for different level of graph connectivity.}
  \label{fig:twobytwo}
\end{figure}

 For most of our results in the Ising section, we pick $k_{\mathrm{con}} = N$ so that the graph is fully connected. This results in the best in-distribution performance. In that case note from Figures \ref{fig:adsgnnfull} and \ref{fig:egnnfull} that AdS-GNN generalizes dramatically better than EGNN. EGNN's performance is particularly bad when taking a model trained on a larger system and evaluating it on a smaller one. 

 However if we reduce $k_{\mathrm{con}}$ -- e.g. to $2$ -- then this greatly increases the generalization ability of both models. In particular, the gap between EGNN and AdS-GNN (as seen in Figures \ref{fig:adsgnn-k-2} and \ref{fig:egnn-k-2}) is much narrower, though still present. We see however that this choice hurts in-distribution performance; AdS-GNN is able to generalize reasonably well even in the fully connected case. 

\subsection{Implementation details} 
In every experiment, we use the AdamW optimizer \cite{LoshchilovH19} with a learning rate $10^{-3}$. Every model is trained on a single Nvidia RTX6000 GPU. All models are implemented in JAX. All experiments are run $5$ times with different seeds. Models are trained until convergence with early stopping. The number of layers is fixed to 4, with hidden dimension 32.

\paragraph{SuperPixel-MNIST} The task is to predict a digit given a point cloud representation. We compare against MONET~\cite{MontiBMRSB17}, SplineCNN~\cite{FeyLWM18}, GCCP~\cite{WalkerG19}, GAT~\cite{VelickovicCCRLB18}, PNCNN~\cite{FinziBW21} and P$\Theta$NITA~\cite{BekkersVHLR24}. Every model is trained with batch size $128$, baseline results are taken from \cite{BekkersVHLR24}. $k_{\text{con}}$ is set to $16$, $k_{\text{lift}}$ to $5$. Training time is approximately 10 minutes.

\paragraph{Pascal-VOC} The task is to predict a semantic segmentation label for each superpixel node (total of $21$ classes). Each graph is embedded in $2$D Euclidean space, each node is associated with $12$ scalar features. We used the batch size of $96$. $k_{\text{con}} = 16$, $k_{\text{lift}} = 5$. Training time is approximately 30 minutes.

\paragraph{Shapes} Given a point cloud, we predict for each point to each shape it belongs (circle, square, triangle, intersection). We train by minimizing cross-entropy loss. $k_{\text{con}}$ is set to $16$, $k_{\text{lift}} = 16$. The number of testing and validation points is set to 512, while the number of training points varies from 64 to 8192. Training time is approximately 5 minutes.

\paragraph{Ising} Given a set of points, we predict two scalar values, one for energy correlation, one for spin correlation. We use relative L2 error as our training objective. $k_{\text{con}}$ is equal to the number of points (i.e. fully connected system), $k_{\text{lift}} = 1$. The number of testing and validation points is set to 512, while the number of training points varies from 64 to 32768. Training time is approximately 3 minutes.

\section{Ising model review} \label{app:ising-review} 
Here, for completeness, we provide an elementary review of the 2d Ising model -- which can be imagined as the simplest example of a model of magnetism -- and the physics at its critical point. This material is standard; see e.g. \cite{kardar2007statistical,DiFrancesco:1997nk} for textbook treatments. 

The model is defined in terms of binary variables called {\it spins} $\sig_i = \pm 1$ sitting on the sites $i$ of a square lattice with $L$ sites on a side. The model is defined in terms of an energy function:
\be
E[\sig] = -\sum_{\vev{ij}} \sig_{i} \sig_{j} \label{ising_en} 
\ee
where the notation $\vev{ij}$ means that one sums over nearest-neighbour links connecting two adjacent sites $i,j$. We can see that the energy is minimized when spins on two adjacent sites have the same value, i.e. ``spins want to align''.  The energy is also invariant under a $\mathbb{Z}_2$ symmetry which acts by flipping all of the spins, $\sig_i \to -\sig_i$. 

This energy defines a statistical physics model in which the probability of obtaining a given spin configuration $\{ \sig \}$ is given by
\be
p_{\beta}[\sig] = \frac{1}{Z} \exp(-\beta E[\sig])
\ee
where $\beta$ is the inverse temperature and $Z$ the usual normalizing constant. This model has two {\it phases}, which we now describe. 

Consider first taking $\beta$ very large; in that case any increase in energy will be heavily penalized, and the most likely configurations will be those that minimize the energy, i.e. where all spins have the same value, so either $\sig_i = +1$ or $\sig_i = -1$ for all $i$. This is called a phase with {\it spontaneous symmetry breaking}, as a choice of either of these configurations breaks the $\mathbb{Z}_2$ symmetry. It is also called the {\it ordered} or {\it ferromagnetic} phase.  

Now consider taking $\beta$ very small; in that case the system is very disordered, and all spins fluctuate strongly and randomly, and there is no sense in which a symmetry is spontaneously broken. This is called the {\it symmetry unbroken} or {\it disordered} phase or the {\it paramagnet.}. 

In the $L \to \infty$ limit there is a sharp distinction between the two phases. A quantitative way to understand it is to imagine taking the system size to infinity while computing e.g. 
\be
\vev{\sig} = \lim_{L \to \infty} \le(\frac{1}{L^2} \mathbb{E}_{\sig \sim p_{\beta}} \sum_{i} \sig_i\ri)
\ee
i.e. the expectation value of the spatial average of all the spins. This is called the order parameter. This is nonzero in the ordered phase (where all the spins are aligned, resulting in a net contribution to the expectation value) and zero in the disordered phase (where all the spins fluctuate strongly, resulting in a cancellation across sites). In the infinite-$L$ limit there is a non-analyticity in the function $\vev{\sig(\beta)}$ at a critical value of $\beta = \beta_{c}$ at the phase transition point. For the 2d Ising model the location of this point is known to be at $\beta_{c} = \ha\log(1+\sqrt{2}) \approx 0.441$. 

A great deal is known about this critical point. Here fluctuations of the spins take place over all scales, and do not decay exponentially with distance as one might normally expect. It is possible to capture the long-distance statistics of these fluctuations in a continuum limit where we formally take the lattice spacing to zero. The resulting structure is called the {\it 2d Ising conformal field theory} (CFT), and is an example of a quantum field theory that exhibits conformal symmetry. In particular, one can define two operators in this continuum theory: the spin operator $\sig(x)$ (which is the continuum limit of the spin operator $\sig_i$ defined on discrete lattice sites above) and the energy operator $\ep(x)$ (which can be thought of as a product of spins at adjacent sites).  The 2d Ising CFT is completely solved and thus one can compute any arbitrary moments of any of these operators in closed form. 

To get some intuition, the correlation function of two spins behaves as: 
\be
\langle \sig(x) \sig(y) \rangle = |x-y|^{-2\Delta_{\sig}}
\ee
where $\Delta_{\sig} = \frac{1}{8}$. The power-law functional form is completely fixed by scale-invariance, and the only input from the theory here is the value of the conformal dimension $\Delta_{\sig}$. A similar relation holds for the energy operator with $\Delta_{\ep} = 1$. 

In this work we build a neural network to predict the $N$-point correlation functions of the spin and energy operators as a function of the positions of the operator insertions. To do this we use as ground-truth training data the following closed-form formulas from the theory of the 2d Ising CFT \cite{DiFrancesco:1997nk}:
\begin{align}
\vev{\ep(z_1) \ep(z_2) \cdots \ep(z_N)}  & = \bigg|\mathrm{Pf}\le[\frac{1}{z_i - z_j}\ri]_{1 < i,j < N}\bigg|^2 \label{pfaffian_app} \\ 
\vev{\sig(z_1) \sig(z_2) \cdots \sig(z_N)}^2  & = \frac{1}{2^{2N}} \sum_{\ep_i = \pm 1, \sum_{\ep_i = 0}} \prod_{i < j} |z_i - z_j|^\frac{\ep_i \ep_j}{2} \label{spincorr-app} 
\end{align} 
where $\mathrm{Pf}$ denotes the matrix Pfaffian. Despite their compact presentation, these are rather complicated functions; e.g. combinatorially the Pfaffian can be viewed as a sum over all possible perfect matchings of $N$ points. The number of these matchings grows factorially in $N$, and each of them contributes a (different) product of 2-body interactions. There are more efficient ways than this to actually {\it compute} the Pfaffian, but our neural networks will be unable to realize that structure. 
\newpage

\input{conformal-theory-script}

\end{document}

%% file: thm-env.tex
\newtheorem{sa}{Satz}[subsection]
\newtheorem{Thm}[sa]{Theorem}
\newtheorem{Lem}[sa]{Lemma}

\newtheorem{Prp}[sa]{Proposition}
\newtheorem{Cor}[sa]{Corollary}

\newtheorem{Def}[sa]{Definition}
\newtheorem{DefLem}[sa]{Definition/Lemma}
\newtheorem{NotLem}[sa]{Notation/Lemma}
\newtheorem{DefsNot}[sa]{Definitions/Notations}

\newtheorem{Not}[sa]{Notation}

\newtheorem{Rem}[sa]{Remark}

\newtheorem{Eg}[sa]{Example}

%% file: math-commands.tex
\def\eqref#1{equation~\ref{#1}}

\def\1{\bm{1}}

\def\ri{{\textnormal{i}}}

\DeclareMathAlphabet{\mathsfit}{\encodingdefault}{\sfdefault}{m}{sl}
\SetMathAlphabet{\mathsfit}{bold}{\encodingdefault}{\sfdefault}{bx}{n}

\def\sO{{\mathbb{O}}}

\DeclareMathOperator{\GConf}{\Conf_g}

\def\({\left(}
\def\){\right)}
\def\[{\left[}
\def\]{\right]}
\def\<{\langle}
\def\>{\rangle}

\newcommand{\bmat}{\begin{bmatrix}}
\newcommand{\emat}{\end{bmatrix}}

\newcommand\half{{\ensuremath{\frac{1}{2}}}}

\newcommand\vev[1]{{\ensuremath{\left\langle{#1}\right\rangle}}}

\newcommand{\be}{\begin{equation}}
\newcommand{\ee}{\end{equation}}
\newcommand{\bea}{\begin{eqnarray}}
\newcommand{\eea}{\end{eqnarray}}
\newcommand{\bwt}{\begin{widetext}}
\newcommand{\ewt}{\end{widetext}}

\newcommand{\bi}{\begin{itemize}}
\newcommand{\ei}{\end{itemize}}
\newcommand{\ben}{\begin{enumerate}}
\newcommand{\een}{\end{enumerate}}
\newcommand{\bca}{\begin{cases}}
\newcommand{\eca}{\end{cases}}
\newcommand{\bln}{\begin{align}}
\newcommand{\eln}{\end{align}}
\newcommand{\bst}{\begin{split}}
\newcommand{\est}{\end{split}}

\newcommand\ep{\epsilon}
\newcommand\sig{\sigma}

\newcommand\lam{\lambda}

\newcommand\ha{{\half}}

\def\le{\left}
\def\ri{\right}

%% file: more-commands.tex
\newcommand{\R}{\mathbb{R}}

\newcommand{\x}{\times}						
\newcommand{\id}{\mathrm{id}}

\newcommand{\Zcal}{\mathcal{Z}}

\newcommand{\Abb}{\mathbb{A}}

\newcommand{\Mbb}{\mathbb{M}}

\newcommand{\Pbb}{\mathbb{P}}

\newcommand{\Rbb}{\mathbb{R}}
\newcommand{\Sbb}{\mathbb{S}}

\newcommand{\sm}{\setminus}						
\newcommand{\ins}{\subseteq}

\newcommand{\sdp}{\rtimes}
\DeclareMathOperator{\sgn}{sgn}

\newcommand{\lp}{\left ( }
\newcommand{\rp}{\right ) }

\newcommand{\lB}{\left [ }
\newcommand{\rB}{\right ] }
\newcommand{\lC}{\left \{ }
\newcommand{\rC}{\right \} }

\newcommand{\st}{\,\middle|\,}

\DeclareMathOperator*{\diag}{\mathrm{diag}}
\newcommand{\dcup}{\,\dot{\cup}\,}

\newcommand*{\matb}[1]{\begin{bmatrix}#1\end{bmatrix}}

\makeatletter
\def\widebreve{\mathpalette\wide@breve}
\def\wide@breve#1#2{\sbox\z@{$#1#2$}%
     \mathop{\vbox{\m@th\ialign{##\crcr
\kern0.08em\brevefill#1{0.8\wd\z@}\crcr\noalign{\nointerlineskip}%
                    $\hss#1#2\hss$\crcr}}}\limits}
\def\brevefill#1#2{$\m@th\sbox\tw@{$#1($}%
  \hss\resizebox{#2}{\wd\tw@}{\rotatebox[origin=c]{90}{\upshape(}}\hss$}
\makeatletter

\DeclareMathOperator{\Isom}{Isom}
\DeclareMathOperator{\Conf}{Conf}
\DeclareMathOperator{\ConfDiff}{ConfDiff}
\DeclareMathOperator{\dS}{dS}
\DeclareMathOperator{\AdS}{AdS}
\DeclareMathOperator{\tAdS}{\widebreve{\AdS}}
\newcommand{\Tan}{\mathrm{T}}
\newcommand{\invsp}{\varsigma}
\DeclareMathOperator{\Or}{O}
\DeclareMathOperator{\SOr}{SO}
\DeclareMathOperator{\GL}{GL}
\DeclareMathOperator{\COr}{CO}
\DeclareMathOperator{\CEu}{CE}
\DeclareMathOperator{\POr}{PO}

\DeclareMathOperator{\arccosh}{arccosh}

%% file: figures/cg_main_fig.tex
\begin{wrapfigure}{R}{0.55\linewidth}
\vspace{-1em}
\centering
        \begin{tikzpicture}
        \node[anchor=south west,inner sep=0] (image) at (0,0) 
            {\includegraphics[width=0.8\linewidth]{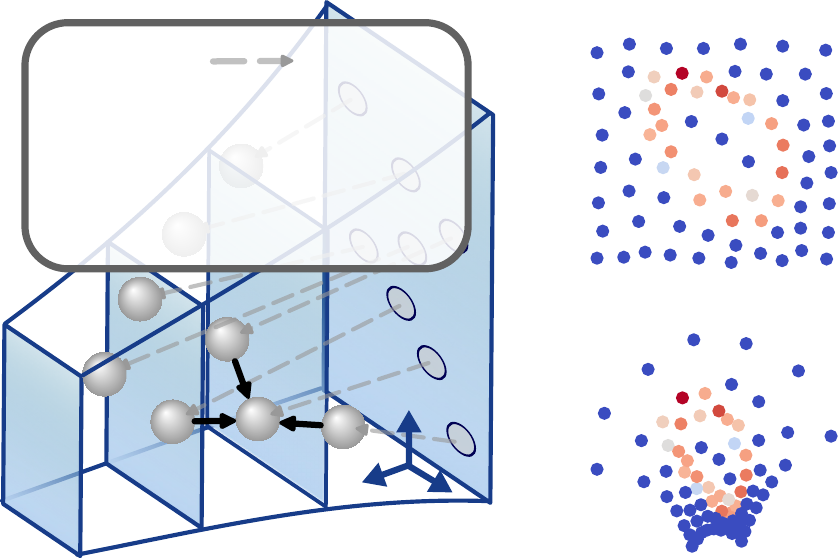}};

        \begin{scope}[x={(image.south east)},y={(image.north west)}]
            \node[color=black, font=\fontsize{8}{8}\selectfont] at (.17,.89) {$\mathbf{x}_i, \mathbf{h}_i$};
            \node[color=black, font=\fontsize{8}{8}\selectfont] at (.44,.89) {$X_i, \mathbf{h}_i$};
            \node[color=black, font=\fontsize{8}{8}\selectfont] at (.29,.8) {\textbf{AdS Embedding}};
        \end{scope}

        \begin{scope}[x={(image.south east)},y={(image.north west)}]
            \node[color=black, font=\fontsize{8}{8}\selectfont] at (.3,.67) {$\psi\left(\mathbf{h}_i, \mathbf{h}_j, D(X_i, X_j)\right)$};
            \node[color=black, font=\fontsize{8}{8}\selectfont] at (.29,.58) {\textbf{AdS-GNN Message}};
        \end{scope}

        \begin{scope}[x={(image.south east)},y={(image.north west)}]
            \node[color=black, rotate=-90] at (.63,0.5) {boundary};
            \node[color=black] at (.53,0.28) {$x^2$};
            \node[color=black] at (.55,0.05) {$x^1$};
            \node[color=black] at (.415,0.13) {$z$};
        \end{scope}

        \begin{scope}[x={(image.south east)},y={(image.north west)}]
            \node[color=black] at (.85,1.) {label: $0$};
            \node[color=black] at (.85,.45) {Conf. transf-n};
        \end{scope}

    \end{tikzpicture}
    \caption{AdS-GNN lifts points from Euclidean space to Anti de Sitter space and computes message passing conditioned on the proper distance.}
    \label{fig:ads}
    \vspace{-30pt}
\end{wrapfigure}

%% file: figures/hyperboloid.tex
\begin{tikzpicture}
    \begin{axis}[
    width=7cm,
    height=7cm,
    view={45}{30},
    axis lines=center,
    xlabel=\empty,
    ylabel=\empty,
    zlabel=\empty,
    xmin=-2.5, xmax=2.5,
    ymin=-2.5, ymax=2.5,
    zmin=-2.5, zmax=2.5,
    xtick=\empty,
    ytick=\empty,
    ztick=\empty,
    grid=none,
    trim axis left,
    clip=false,
]
\addplot3[
    mesh,
    domain=-2:2,
    domain y=0:2*pi,
    samples=15,
    samples y=20,
    black!50,
    no marks,
    z buffer=sort,
]
({x*cos(deg(y))}, {x*sin(deg(y))}, {sqrt(1+x^2)});
\addplot3[
    mesh,
    domain=-2:2,
    domain y=0:2*pi,
    samples=15,
    samples y=20,
    black!50,
    no marks,
    z buffer=sort,
]
({x*cos(deg(y))}, {x*sin(deg(y))}, {-sqrt(1+x^2)});
\foreach \r in {0.85,1.42,2}{
    \addplot3[
        domain=0:2*pi,
        samples=30,
        black!70,
        thick,
    ]
    ({\r*cos(deg(x))}, {\r*sin(deg(x))}, {sqrt(1+\r^2)});
}
\foreach \r in {0.85,1.42,2}{
    \addplot3[
        domain=0:2*pi,
        samples=30,
        black!70,
        thick,
    ]
    ({\r*cos(deg(x))}, {\r*sin(deg(x))}, {-sqrt(1+\r^2)});
}
\node[fill=white, fill opacity=0.7, text opacity=1] at (axis cs:0,0,3) {$Y^{0}$};
\node[fill=white, fill opacity=0.7, text opacity=1] at (axis cs:0,3,0) {$Y^{2}$};
\node[fill=white, fill opacity=0.7, text opacity=1] at (axis cs:3,0,0) {$Y^{1}$};
\end{axis}
\end{tikzpicture}

%% file: figures/lifting.tex
\begin{tikzpicture}
    \node[anchor=south west,inner sep=0] (image) at (0,0) 
        {\includegraphics[width=\linewidth]{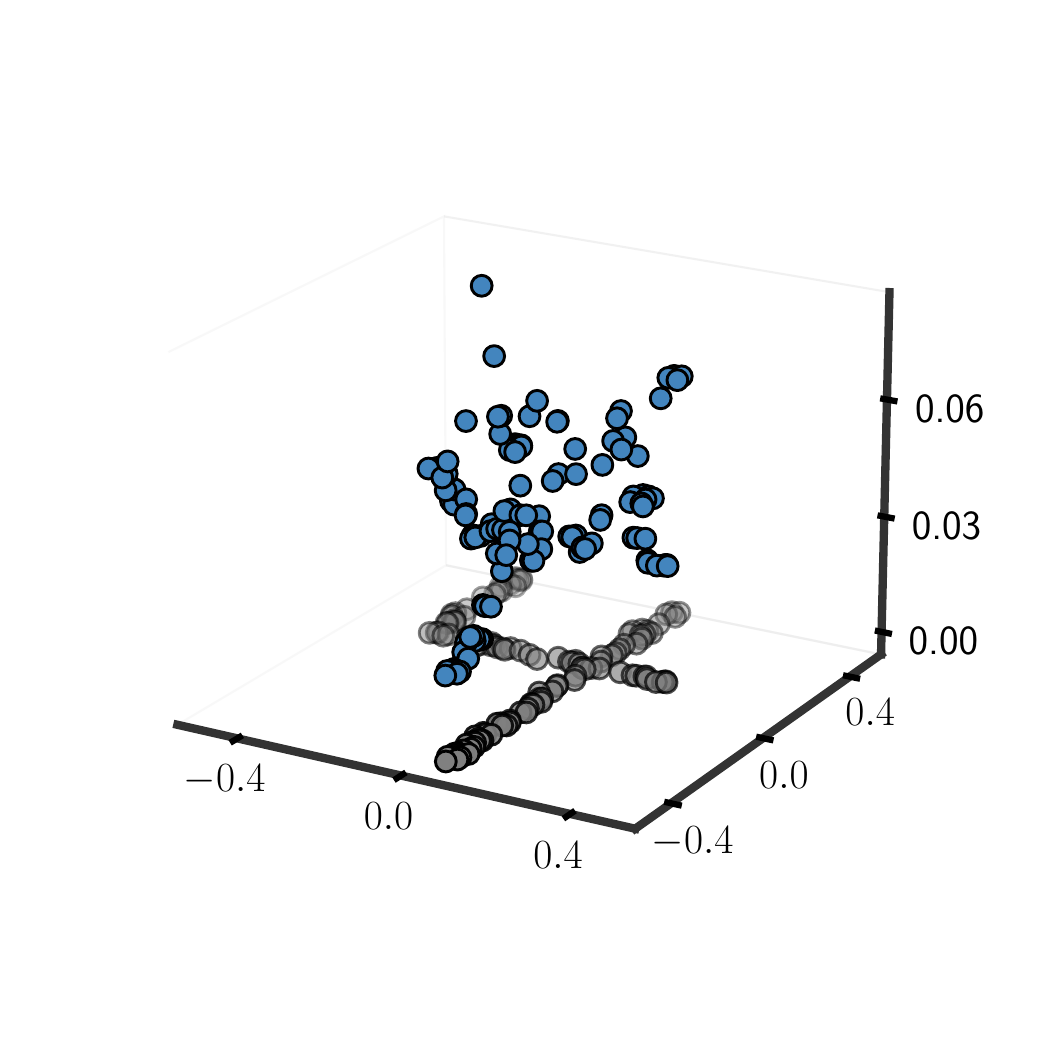}};

    \begin{scope}[x={(image.south east)},y={(image.north west)}]
        \node[color=black, font=\fontsize{9}{9}\selectfont] at (0.3,0.175) {$x^1$};
    \end{scope}

    \begin{scope}[x={(image.south east)},y={(image.north west)}]
        \node[color=black, font=\fontsize{9}{9}\selectfont] at (0.85,0.225) {$x^2$};
    \end{scope}

    \begin{scope}[x={(image.south east)},y={(image.north west)}]
        \node[color=black, font=\fontsize{9}{9}\selectfont] at (0.9,0.725) {$z$};
    \end{scope}

\end{tikzpicture}

%% file: conformal-theory-script.tex
\section{The Conformal Group of the Pseudo-Euclidean Space} \label{app:conf} 

In this section we will provide a self-contained introduction to conformal geometry of the (pseudo-) Euclidean space in arbitrary signature and its conformal group. We provide definitions, examples and main theorems.

We follow the references of \cite{Scho08} and \cite{McK25}.

\subsection{Conformal Transformations of the Pseudo-Euclidean Space}

Roughly speaking, a conformal map is a map that preserves local angles. However, this definition is rather tedious to write down and work with. A technically more convenient definition is introduced below in \Cref{def:conf-map}. To motivate this definition, we first show in \Cref{lem:conf-angle-pres} that in the (linear) Euclidean case these two definitions are equivalent:

\begin{Lem}
    \label{lem:conf-angle-pres}
    Let $A \in \R^{d \times d}$ be a real \emph{invertible} $(d\times d)$-matrix.
    Then the following statements are equivalent:
    \begin{enumerate}
        \item $A$ is \emph{angle preserving}, i.e.\ for all $v_1,v_2 \in \R^d\sm\{0\}$ we have:
            \begin{align}
                \frac{\langle Av_1,Av_2\rangle}{\|Av_1\|\cdot\|Av_2\|} &= \frac{\langle v_1,v_2\rangle}{\|v_1\|\cdot\|v_2\|}. 
            \end{align}
        \item $A$ is a \emph{conformal matrix}, i.e.\ $A =c \Lambda$ with a scalar $c >0 $ and an orthogonal matrix $\Lambda \in \Or(d)$.
    \end{enumerate}
    \begin{proof}
        ``$\Longleftarrow$'': Let $A=c\Lambda$ with $c>0$ and $\Lambda\in\Or(d)$.
        Then we get for all $v_1,v_2 \in \R^d$:
        \begin{align}
            \langle Av_1,Av_2\rangle &= v_1^\top A^\top A v_2 \\
                                     &= c^2 \cdot v_1^\top \underbrace{\Lambda^\top \Lambda}_{=I} v_2 \\
                                     &= c^2 \cdot v_1^\top v_2 \\
                                     &= c^2 \cdot \langle v_1,v_2\rangle.
        \end{align}
        This also implies:
        \begin{align}
            \|Av_1\| &= c \cdot \|v_1\|, & \|Av_2\| &= c \cdot \|v_2\|.
        \end{align}
        Together this implies the claim:
            \begin{align}
                \frac{\langle Av_1,Av_2\rangle}{\|Av_1\|\cdot\|Av_2\|} &= \frac{\langle v_1,v_2\rangle}{\|v_1\|\cdot\|v_2\|}. 
            \end{align}
            ``$\Longrightarrow$'': Assume that $A$ is invertible and preserves angles. Then $C:=A^\top A$ is symmetric and positive definite. By the spectral theorem we can diagonalize $C$, i.e.\ there exists an orthonormal basis $e_1,\dots,e_d \in \R^d$, $\langle e_i, e_j \rangle = \delta_{i,j}$ for all $i,j \in [d]$, and positive scalars $\lambda_1,\dots, \lambda_d > 0$ such that for all $i \in [d]$:
            \begin{align}
                C e_i & = \lambda_i\cdot e_i.
            \end{align}
     We now claim that: $\lambda_1 = \cdots = \lambda_d$.
     By way of contradiction assume that there exists $\ell \in [d]$ such that $\lambda_1 \neq \lambda_\ell$. Then put: $v_1:=e_1$ and $v_2 := e_1 + e_\ell$. This gives:
     \begin{align}
         \|v_1\|^2 &= 1, &  \|v_2\|^2 &= \|e_1\|^2+\|e_\ell\|^2 = 2, \\
         \|Av_1\|^2 &= \lambda_1, & A^\top Av_2 &= \lambda_1\cdot e_1 + \lambda_\ell \cdot e_\ell, \\
         \langle Av_1,Av_2\rangle &= \lambda_1 ,&  \|Av_2\|^2 &= \lambda_1+\lambda_2,\\
         \langle v_1,v_2\rangle &= 1.
     \end{align}
     This implies:
            \begin{align}
                \frac{\lambda_1}{\sqrt{\lambda_1}\cdot \sqrt{\lambda_1+\lambda_\ell}}   =    \frac{\langle Av_1,Av_2\rangle}{\|Av_1\|\cdot\|Av_2\|} &= \frac{\langle v_1,v_2\rangle}{\|v_1\|\cdot\|v_2\|} = \frac{1}{1 \cdot \sqrt{2}}. 
            \end{align}
    Squaring and solving for $\lambda_\ell$ shows: 
    \begin{align}
        \lambda_\ell &=\lambda_1,
    \end{align}
    which contradicts the assumption. So, we indeed have that:
    \begin{align}
        c^2 := \lambda_1 =\cdots= \lambda_d > 0.
    \end{align}
    This shows that: $C= c^2 I$. Putting $\Lambda := \frac{1}{c}A$ we get:
    \begin{align}
        \Lambda ^\top \Lambda = \frac{1}{c^2}C = I,
    \end{align}
    which shows that $\Lambda \in \Or(d)$ and thus $A =c \Lambda$ with $c >0$ and $\Lambda \in \Or(d)$.
    \end{proof}
\end{Lem}

\begin{Def}[Conformal maps/conformal transformations]
    \label{def:conf-map}
    Let $(M,\eta^M)$ and $(N,\eta^N)$ be two pseudo-Riemannian manifolds.
    A \emph{conformal map} $f: M \to N$ is defined to be a smooth map such that there exists a smooth map $\omega:M \to \R_{>0}$ such that for all $x \in M$ and $v_1,v_2 \in \Tan_xM$ we have:
    \begin{align}
        \eta^N_{f(x)}(df_x(v_1),df_x(v_2)) &= \omega(x)^2 \cdot \eta^M_x(v_1,v_2),
    \end{align}
    where $df_x:\Tan_xM \to \Tan_{f(x)}N$ is the differential of $f$ at $x \in M$.
    The above map $\omega$ is called the \emph{conformal factor} of $f$.

    In case the conformal factor $\omega$ of $f$ equals the constant one, $\omega(x)=1$, then we call $f$ an \emph{isometric map} or isometric transformation.
\end{Def}

\begin{Def}[Conformal diffeomorphisms and isometries]
     Let $(M,\eta^M)$ be a pseudo-Riemannian manifold.
     A \emph{conformal diffeomorphism} $f:M \to M$ is a conformal smooth map that has a conformal smooth inverse $f^{-1}: M \to M$: 
     \begin{align} f\circ f^{-1} = f^{-1} \circ f = \id_M. \end{align}
     If, in addition, its conformal factor $\omega$ of $f$ is the contant $1$, then we call $f$ an \emph{isometry} of $(M,\eta^M)$.
        The \emph{group of conformal diffeomorphisms} of $(M,\eta^M)$ is denoted as:
    \begin{align}
        \ConfDiff(M,\eta^M) &:= \lC f: M \to M \text{ conformal map with conformal inverse} \rC.
    \end{align}
        The \emph{isometry group} of $(M,\eta^M)$ is denoted as:
    \begin{align}
        \Isom(M,\eta^M) &:= \lC f: M \to M \text{ isometric map with isometric inverse} \rC.
    \end{align}
\end{Def}

\begin{Not}
    In the following we will denote by:
    \begin{enumerate}
        \item $(\R^{p,q},\eta^{p,q})$ be the \emph{standard pseudo-Euclidean space} of signature $(p,q)$:
            \begin{align}
                \eta^{p,q}(v_1,v_2) &:= v_1^\top \Delta^{p,q} v_2, & 
                \Delta^{p,q}:=\diag(\underbrace{+1,\dots,+1}_{\times p},\underbrace{-1,\dots,-1}_{\times q}),
            \end{align}
        \item $\Or(p,q)$ the \emph{(pseudo-)orthogonal group} of signature $(p,q)$:
            \begin{align}
                \Or(p,q) &:= \lC \Lambda \in \GL(\R^{p,q}) \st \Lambda^\top \Delta^{p,q} \Lambda = \Delta^{p,q} \rC,
            \end{align}
        \item $\SOr(p,q)$ the \emph{special (pseudo-)orthogonal group} of signature $(p,q)$:
            \begin{align}
                \SOr(p,q) &:= \lC \Lambda \in \Or(p,q) \st \det \Lambda = 1 \rC.
            \end{align}
        \item $\Or^0(p,q) = \SOr^0(p,q)$ the identity component (=connected component of the identity) of $\Or(p,q)$.
            Note that we have the inclusions of groups:
            \begin{align}
                \Or^0(p,q) \ins \SOr(p,q) \ins \Or(p,q).
            \end{align}
    \end{enumerate}
\end{Not}

\begin{Eg}[Affine conformal diffeomorphisms]
    \label{eg:aff-conf-diff}
  Let $b \in \R^{p,q}$, $c > 0$ and $\Lambda \in \Or(p,q)$. Consider the affine map:
    \begin{align}
        f: \R^{p,q} &\to \R^{p,q}, & f(x) &:= c \Lambda x +b.
    \end{align}
    Then $f$ is a conformal diffeomorphism of $\R^{p,q}$ with constant conformal factor $\omega(x)=c$.
    \begin{proof}
        The differential at $x$ is given by the matrix:
        \begin{align}
            df_x &= c\Lambda.
        \end{align}
        So for $v_1, v_2 \in \Tan_x\R^{p,q}=\R^{p,q}$ we get:
    \begin{align}
        \eta^{p,q}(df_x(v_1),df_x(v_2)) &= (c\Lambda v_1)^\top \Delta^{p,q} (c\Lambda v_2) \\
                                        &= c^2 \cdot v_1^\top \Lambda^\top \Delta^{p,q} \Lambda v_2 \\
                                        &= c^2 \cdot v_1^\top  \Delta^{p,q} v_2 \\
                                        &= c^2 \cdot \eta^{p,q}(v_1, v_2).
    \end{align}
    So, if we then define the conformal factor $\omega: \R^{p,q} \to \R_{>0}$ to be the constant map $\omega(x):=c$, we have shown that $f$ is a conformal smooth map.
    Its inverse $f^{-1}$ is given by:
    \begin{align}
        f^{-1}(x) &:= c^{-1}\Lambda^{-1}x - c^{-1}\Lambda^{-1}b,
    \end{align}
    which is thus of the same form as $f$ and thus also a conformal smooth map. This shows the claim.
    \end{proof}
\end{Eg}

\begin{Thm}[Affine conformal diffeomorphisms, see \cite{Ami67}]
    \label{thm:aff-conf-diff}
    Consider the affine map on $\R^{p,q}$:
    \begin{align}
        f: \R^{p,q} &\to \R^{p,q}, & f(x) &:= Ax +b,
    \end{align}
    with a square matrix $A$ and translation vector $b \in \R^{p,q}$.
    Then $f$ is a conformal map (w.r.t.\ $\eta^{p,q}$) iff there exists a $c \in \R_{>0}$ and a $\Lambda \in \Or(p,q)$ such that $A= c\Lambda$. If this is the case, $f$ is a conformal diffeomorphism and both $c > 0$ and $\Lambda \in \Or(p,q)$ are uniquely determined by $A$ as: $c = \sqrt[d]{|\det A|}$ with $d:=p+q$, and, $\Lambda = \frac{1}{c} A$.
    \begin{proof}
        One direction is proven in \Cref{eg:aff-conf-diff}. 
        For the other direction assume that $f$ is a conformal map. Then we have $df_x = A$ and the conformal relation (in matrix form):
        \begin{align}
            A^\top \Delta^{p,q} A & = \omega(x)^2 \cdot \Delta^{p,q}. 
        \end{align}
        Taking determinants on both sides gives:
        \begin{align}
            (\det A)^2 & = (\omega(x)^2)^d > 0,
        \end{align}
        showing that $A$ is invertible and that $\omega(x) = \sqrt[d]{|\det A|}=:c$ is not dependent on $x$ and thus equal to the constant $c > 0$.
        Dividing the first equation on both sides by $c^2$ and rearranging gives:
        \begin{align}
            \lp \frac{1}{c} A  \rp^\top \Delta^{p,q}  \lp \frac{1}{c} A \rp & = \Delta^{p,q}. 
        \end{align}
        This shows that $\Lambda:=\frac{1}{c} A  \in \Or(p,q)$ and thus the claim: $A=c \Lambda$.
        The rest follows from \Cref{eg:aff-conf-diff}.
    \end{proof}
\end{Thm}

\begin{Def}[The linear and affine conformal group]
    We define the \emph{linear and affine conformal group}, resp., of signature $(p,q)$ as follows:
    \begin{align}
        \COr(p,q) &:= \R_{>0} \times \Or(p,q), & \CEu(p,q) &:= \COr(p,q) \sdp \R^{p,q}.
    \end{align}
Note that the entries correspond to scaling factor $c >0$, reflection-rotation matrix $\Lambda \in \Or(p,q)$ and translation vector $b \in \R^{p,q}$ in the conformal affine maps from \Cref{eg:aff-conf-diff} and \Cref{thm:aff-conf-diff}.

We also define their identity components:
    \begin{align}
        \COr^0(p,q) &:= \R_{>0} \times \Or^0(p,q), & \CEu^0(p,q) &:= \COr^0(p,q) \sdp \R^{p,q}.
    \end{align}
\end{Def}

\begin{Eg}[The inversion at the pseudo-sphere]
    \label{eg:inversion}
    The following (partial) map:
    \begin{align}
        \invsp=\invsp^{p,q}: \R^{p,q} &\to \R^{p,q}, & \invsp^{p,q}(x) &:= \frac{x}{\eta^{p,q}(x,x)},
    \end{align}
    is called the \emph{inversion (at the pseudo-sphere)} of signature $(p,q)$.
    Note that the inversion here is only defined for $x \in U:=\lC \tilde x\in \R^{p,q} \st \eta^{p,q}(\tilde x,\tilde x) \neq 0 \rC$.
    It is an involution (self-invers) on $U$ as (with $\eta:=\eta^{p,q}$):
    \begin{align}
        \eta(\invsp(x),\invsp(x)) &= \frac{\eta(x,x)}{\eta(x,x)^2} = \frac{1}{\eta(x,x)} \neq 0, \\
        \invsp(\invsp(x)) & = \frac{\invsp(x)}{\eta(\invsp(x),\invsp(x))} 
        = \frac{\frac{x}{\eta(x,x)}}{\frac{1}{\eta(x,x)}} = x.
    \end{align}
    We now claim that $\invsp : U \to U$ is a conformal diffeomorphism. 
    For this we compute its Jacobian matrix (differential), which, in matrix form, is given as:
    \begin{align}
        d\invsp_x & = \frac{1}{\eta(x,x)^2}\lp \eta(x,x) I - 2xx^\top \Delta^{p,q} \rp.
    \end{align}
    To check the conformal relation we thus compute:
    \begin{align}
        (d\invsp_x)^\top \Delta^{p,q} (d\invsp_x) 
        &= \frac{1}{\eta(x,x)^4} \lp \lp \eta(x,x) I - 2xx^\top \Delta^{p,q} \rp^\top \Delta^{p,q} \lp \eta(x,x) I - 2xx^\top \Delta^{p,q} \rp \rp \\
        &= \frac{1}{\eta(x,x)^4} \lp \lp \eta(x,x) I - 2 \Delta^{p,q}xx^\top  \rp \Delta^{p,q} \lp \eta(x,x) I - 2xx^\top \Delta^{p,q} \rp \rp \\
        &= \frac{1}{\eta(x,x)^4} \lp \eta(x,x)^2 \Delta^{p,q}  - 4 \eta(x,x)\Delta^{p,q}xx^\top\Delta^{p,q} + 4 \Delta^{p,q} xx^\top \Delta^{p,q} xx^\top \Delta^{p,q} \rp \\
        &= \frac{1}{\eta(x,x)^4} \lp \eta(x,x)^2 \Delta^{p,q}  - 4 \eta(x,x)\Delta^{p,q}xx^\top\Delta^{p,q} + 4 \eta(x,x)\Delta^{p,q} xx^\top \Delta^{p,q} \rp \\
        &= \frac{1}{\eta(x,x)^2}\Delta^{p,q}.
    \end{align}
    This shows that the inversion $\invsp:U \to U$ is a conformal diffeomorphism with conformal factor $\omega(x)=\frac{1}{|\eta(x,x)|}$.
    We continue the analysis in \Cref{eg:inversion-2}.
\end{Eg}

\begin{Eg}[Special conformal transformations]
    \label{eg:spec-conf-trafo}
    For $b \in \R^{p,q}$ we define the (partial) map:
    \begin{align}
        \sigma_b: \R^{p,q} &\to \R^{p,q}, & 
        \sigma_b(x) &:=  \frac{x - \eta(x,x)\cdot b}{1-2 \cdot \eta(x,b) + \eta(b,b)\cdot\eta(x,x)} = \invsp\lp \invsp(x) - b \rp,
    \end{align}
    where $\invsp=\invsp^{p,q}: \R^{p,q} \to \R^{p,q}$ denotes the inversion from \Cref{eg:inversion} of signature $(p,q)$. Maps of the form $\sigma_b$ are called \emph{special conformal transformations} of $\R^{p,q}$ and are conformal maps on their domain of definition: 
    \begin{align} 
        U_b := \lC x \in \R^{p,q} \st \nu(x,b) \neq 0 \rC,
    \end{align}
    with conformal factor $\omega_b(x) := \frac{1}{|\nu(x,b)|}$,
    where we abbreviated the above denominator as:
    \begin{align}
        \nu(x,b) &:= 1-2 \cdot \eta(x,b) + \eta(b,b)\cdot\eta(x,x).
    \end{align}
    Furthermore, $\sigma_{-b}:U_{-b} \to U_b$ is the inverse of $\sigma_b: U_b \to U_{-b}$.
\begin{proof}
    We first check that $y := \sigma_b(x)$ lies in $U_{-b}$ for $x \in U_b$.
    For this we compute:
    \begin{align}
        \eta(x-\eta(x,x)b,x-\eta(x,x)b) &= \eta(x,x) - 2 \eta(x,x) \eta(x,b) + \eta(x,x)^2 \eta(b,b) \\
                                        &= \eta(x,x) \nu(x,b), \\
        \eta(x-\eta(x,x)b,-b) &= -\eta(x,b) + \eta(x,x)\eta(b,b)
    \end{align}
    Dividing both sides by $\nu(x,b)^2$ and $\nu(x,b)$, resp.,  shows:
    \begin{align}
        \eta(y,y) &= \frac{\eta(x,x)}{\nu(x,b)}, & \eta(y,-b) &= \frac{1}{\nu(x,b)} \lp -\eta(x,b) + \eta(x,x)\eta(b,b) \rp. \label{eq:spec-conf-trafo-rel}
    \end{align}
    With this we get:
    \begin{align}
        \nu(y,-b) & = 1-2 \cdot \eta(y,-b) + \eta(b,b)\cdot\eta(y,y) \\
                  &= \frac{1}{\nu(x,b)} \lp \nu(x,b) + 2 \cdot \eta(x,b) -2 \cdot \eta(x,x)\cdot \eta(b,b) + \eta(b,b) \cdot \eta(x,x) \rp \\
                  &= \frac{1}{\nu(x,b)} \lp \nu(x,b) + 2 \cdot \eta(x,b) - \eta(b,b) \cdot \eta(x,x) \rp \\
                  &= \frac{1}{\nu(x,b)} \\
                  & \neq 0.
    \end{align}
    This shows: $y \in U_{-b}$ for $x \in U_b$. Plugging the relation $\nu(y,-b)=\frac{1}{\nu(x,b)}$ into \Cref{eq:spec-conf-trafo-rel} shows that:
    \begin{align}
        \sigma_{-b}(y) &= \frac{y + \eta(y,y) \cdot b}{\nu(y,-b)} \\
        &=\frac{1}{\nu(y,-b)}\cdot y + \frac{\eta(y,y)}{\nu(y,-b)} \cdot b \\
        &=\nu(x,b) \cdot \lp \frac{x - \eta(x,x) \cdot b}{\nu(x,b)}  \rp + \eta(x,x) \cdot b \\ 
        & = x.
    \end{align}
    This shows that $\sigma_{-b}:U_{-b} \to U_b$ is the inverse of $\sigma_b: U_b \to U_{-b}$.

    To show that $\sigma_b: U_b \to \R^{p,q}$ is a conformal map we need to compute its differential. 
    Computing the differential directly gives:
    \begin{align}
        (d\sigma_b)_x &= \frac{\nu(x,b) \cdot \lp I - 2\cdot b \cdot x^\top \Delta^{p,q} \rp - \lp x - \eta(x,x) \cdot b \rp \cdot \lp -2 \cdot b^\top \Delta^{p,q} + 2 \cdot \eta(b,b) \cdot x^\top \Delta^{p,q} \rp }{\nu(x,b)^2},
    \end{align}
    which is difficult to work with. Instead, we use the fact that $\sigma_b$ is a smooth extension of the map $\sigma_b(x) = \invsp\lp\invsp(x) -b \rp$. With the chain rule we then get:
    \begin{align}
        (d\sigma_b)_x &= (d\invsp)_{(\invsp(x)-b)}\, d\invsp_x
    \end{align}
    With this and \Cref{eg:inversion} we get:
    \begin{align}
        (d\sigma_b)_x^\top \Delta^{p,q} (d\sigma_b)_x 
        & = (d\invsp_x)^\top \, (d\invsp)_{(\invsp(x)-b)}^\top \,  \Delta^{p,q} \,(d\invsp)_{(\invsp(x)-b)}\, d\invsp_x \\
        & = \frac{1}{\eta\lp \invsp(x)-b, \invsp(x)-b \rp^2 } \cdot (d\invsp_x)^\top \,   \Delta^{p,q} \, d\invsp_x \\
        &= \frac{1}{\eta(x,x)^2}\cdot \frac{1}{\eta\lp \invsp(x)-b, \invsp(x)-b \rp^2 }\Delta^{p,q} \\
        &= \frac{1}{\nu(x,b)^2 }\Delta^{p,q}.
    \end{align}
    For the last step note that:
    \begin{align}
        \eta\lp \invsp(x)-b, \invsp(x)-b \rp & = \frac{\nu(x,b)}{\eta(x,x)}.
    \end{align}
    So the conformal factor of $\sigma_b$ at $x \in U_b$ is $\omega_b(x):= \frac{1}{|\nu(x,b)|}$. 
    We continue the analysis in \Cref{eg:spec-conf-trafo-2}.
\end{proof}
\end{Eg}

\begin{Rem}[Extending conformal maps to a conformal compactification]
    \label{rem:conf-comp}
    The examples \ref{eg:aff-conf-diff}, \ref{thm:aff-conf-diff}, \ref{eg:inversion}, \ref{eg:spec-conf-trafo} have given us several conformal transformations of $\R^{p,q}$. However, for us to be able to define the inversion $\invsp: U \to \R^{p,q}$ and the special conformal transformations $\sigma_b: U_b \to \R^{p,q}$ we had to restrict their domain of definition to an open subset of $\R^{p,q}$.
    The question is now if we can actually extend those maps like $\invsp$ and $\sigma_b$ to a conformal map on the whole of $\R^{p,q}$ by possibly enlarging their codomain, e.g.\ by putting:
    \begin{align}
        \invsp: \R^{p,q} &\to \R^{p,q} \cup \lC \infty \rC, & \invsp(x) &:= 
        \begin{cases}
            \frac{x}{\eta(x,x)}, & \text{ if } \eta(x,x) \neq 0,\\
            \infty, & \text{ if } \eta(x,x) = 0\,?
        \end{cases}
    \end{align}
 However, for this we would need to turn the extended codomain $\R^{p,q} \cup \lC \infty \rC$ into a proper pseudo-Riemannian manifold $\Mbb^{p,q}$. 
 Furthermore, we then would also like to properly define $\invsp$ on the added new points $\Mbb^{p,q}\sm \R^{p,q}$ and turn it into a conformal diffeomorphism of $\Mbb^{p,q}$. Note that in the non-Euclidean case, i.e.\ if $p,q > 0$, the space $\lC x \in \R^{p,q} \st \eta(x,x) = 0 \rC$ consists of more than one point and thus needs more consideration.

 The above is the main question (of the existence) of a \emph{conformal compactification} of $\R^{p,q}$.
 Below we follow an ad hoc approach to this question.
\end{Rem}

\begin{Rem}
    Even after the question of conformal compactification, as described in \Cref{rem:conf-comp}, is solved, we are still faced with the ambiguity of how to define the \emph{conformal group} of $\R^{p,q}$. There are several non-equivalent options:
    \begin{enumerate}
        \item as the group $\ConfDiff(\R^{p,q})$ of all conformal diffeomorphisms of $\R^{p,q}$, which would include all affine conformal transformations, but exclude the inversion and the special conformal transformations;
        \item as the group $\ConfDiff(\Mbb^{p,q})$ of all conformal diffeomorphisms of a conformal compactification $\Mbb^{p,q}$ of $\R^{p,q}$, which would include all affine and special conformal transformations and the inversion. However, one can show that for $p+q=2$ some of the properties of the conformal compactification will break down and that $\ConfDiff(\Mbb^{p,q})$ can become pathologically big in the case $(p,q)=(1,1)$, see \cite{Scho08};
        \item as the connected component of the identity of one of the above groups, in order to stick to the main conformal diffeomorphisms that have descriptions with help of a Lie algebra and Lie exponential map, etc.;
        \item as the subgroup of (partially defined) conformal diffeomorphisms of $\R^{p,q}$ that is generated by all affine and special conformal transformations (excluding the inversion) of $\R^{p,q}$;
        \item as the subgroup of conformal diffeomorphisms of $\Mbb^{p,q}$ that is generated by all affine and special conformal transformations (excluding the inversion) of $\R^{p,q}$;
    \end{enumerate}
    For this draft we will settle with one of the last definitions, which are mostly equivalent, and call it the \emph{(restricted) conformal group of $\R^{p,q}$} and denote it by $\Conf(\R^{p,q})$. For more details see further below.
\end{Rem}

To define the conformal compactification of $\R^{p,q}$ we first need to introduce the bigger pseudo-Euclidean space $\R^{p+1,q+1}$ and the projective space $\Pbb^{p+q+1}$.

\subsection{The Projective Space, de Sitter Space and Anti-de Sitter Space}

\begin{Def}[The projective space]
    Let $V$ be a (real) finite dimensional vector space. 
    We then define the \emph{projective space associated with $V$} as the space of equivalence classes:
    \begin{align}
        \Pbb(V) &:= \lp V\sm\{0\} \rp/\mathord{\sim},
    \end{align}
    where we define two non-zero vectors $x_1,x_2 \in V$ to be equivalent if they lie on the same straight line through the origin:
    \begin{align}
        x_1 \sim x_2 &:\iff \exists c \in \R \sm \{0\}. \; x_1 = c \cdot x_2.
    \end{align}
    We then define the \emph{standard projective space} of dimensions $d \ge 0$ as:
    \begin{align}
        \Pbb^d &:= \Pbb(\R^{d+1}).
    \end{align}
    Note that $d$ denotes the dimension of $\Pbb^d$, which thus requires the Euclidean space $\R^{d+1}$ to be of one dimension higher.
\end{Def}

\begin{Rem}
    W.r.t.\ to our pseudo-Euclidean space $\R^{p,q}$ we will in the following mostly consider the projective space of dimension $d+1$, where $d:=p+q$:
    \begin{align}
        \Pbb^{d+1} &:= \Pbb(\R^{p+1,q+1}).
    \end{align}
    Elements of $\Pbb^{d+1}$ will be either denoted by $[z]$ with $z \in \R^{p+1,q+1}$ with coordinates $z=(z^0,z^1,\dots,z^d,z^{d+1})$ or directly as $[z]=[z^0:z^1:\dots:z^d:z^{d+1}]$.
    Note that we endow $\R^{p+1,q+1}$ with the standard metric $\eta^{p+1,q+1}$ of the pseudo-Euclidean space of dimension $d+2$ and signature $(p+1,q+1)$ and that we consider the coordinates $z^0,\dots, z^p$ to belong to the $(+1)$-signature and the coordinates $z^{p+1},\dots,z^{d+1}$ to belong to the $(-1)$-signature.
\end{Rem}

\begin{Def}[de Sitter and anti-de Sitter space]
    \label{def:dS-AdS}
    Let $(V,\eta)$ be a non-degenerate finite dimensional real quadratic vector space.
    Then we define the \emph{de Sitter space} and the \emph{anti-de Sitter space} and the \emph{(projective) zero quadric} associated to $(V,\eta)$ as:
    \begin{align}
        \dS(V,\eta) &:= \lC [z] \in \Pbb(V) \st \eta(z,z) > 0 \rC, \\
        \AdS(V,\eta) &:= \lC [z] \in \Pbb(V) \st \eta(z,z) < 0 \rC. \\
        \Mbb(V,\eta) &:= \lC [z] \in \Pbb(V) \st \eta(z,z) = 0 \rC. 
    \end{align}
    Note that these are all well-defined subsets of $\Pbb(V)$.
    The pseudo-Riemannian metrics on the first two spaces are induced by the following subspaces of $V$:
    \begin{align}
        Y_+ &:= \lC y \in V \st \eta(y,y) = 1 \rC, &
        Y_- &:= \lC y \in V \st \eta(y,y) = -1 \rC,
    \end{align}
    and the $2:1$ locally diffeomorphic surjective maps:
    \begin{align}
        \pi_+: Y_+ &\to \dS(V,\eta) \ins \Pbb(V), & \pm y &\mapsto [y], \\
        \pi_-: Y_- &\to \AdS(V,\eta) \ins \Pbb(V), & \pm y &\mapsto [y],
    \end{align}
    where both $Y_+$ and $Y_-$ are endowed with the pull-back metrics $\eta^+$ and $\eta^-$ from $\eta$ of $V$ to those subspaces. More explicitely, the tangent spaces and local metrics are given as follows:
    \begin{align}
        \Tan_{[z]}\dS(V,\eta) &= \lC v \in V \st \eta(z,v) = 0 \rC, & \eta^{\dS(V,\eta)}_{[z]}(v_1,v_2) &= \eta(v_1,v_2), \\
        \Tan_{[z]}\AdS(V,\eta) &= \lC v \in V \st \eta(z,v) = 0 \rC, & \eta^{\AdS(V,\eta)}_{[z]}(v_1,v_2) &= \eta(v_1,v_2).
    \end{align}
    We define the \emph{(standard) de Sitter space} and the \emph{(standard) anti-de Sitter space} and \emph{(standard) zero quadric} of signatures $(p,q)$, resp., as follows:
    \begin{align}
        \dS^{p,q} &:= \dS(\R^{p+1,q},\eta^{p+1,q}), \\
        \AdS^{p,q} &:= \AdS(\R^{p,q+1},\eta^{p,q+1}), \\
        \Mbb^{p,q} &:= \Mbb(\R^{p+1,q+1},\eta^{p+1,q+1}).
    \end{align}
    The pseudo-Riemannian metric on $\Mbb^{p,q}$ will be defined later in \Cref{def:conf-comp-metric}.
\end{Def}

\begin{Def}[Projective orthogonal group]
    Let $(V,\eta)$ be a non-degenerate finite dimensional real quadratic vector space.
    Then the \emph{projective orthogonal group} of $(V,\eta)$ is defined to be:
    \begin{align}
        \POr(V,\eta) &:= \Or(V,\eta)/\{\pm \id_V\}. 
    \end{align}
    The \emph{(standard) projective orthogonal group of signature $(p,q)$} is:
    \begin{align}
        \POr(p,q) &:= \Or(p,q)/\{\pm I\}. 
    \end{align}
    We denote the identity component of those groups as:
    \begin{align}
        \POr^0(V,\eta), \qquad \text{ and } \qquad \POr^0(p,q).
    \end{align}
\end{Def}

\begin{Lem}
  Let $(V,\eta)$ be a non-degenerate finite dimensional real quadratic vector space.
  Then the projective orthogonal group $\POr(V,\eta)$ acts on the spaces $\Pbb(V,\eta)$, $\dS(V,\eta)$, $\AdS(V,\eta)$ and on $\Mbb(V,\eta)$  via matrix multiplication:
  \begin{align}
      [\Lambda][z] &:=[ \Lambda z],
  \end{align}
  in a well-defined way.
\end{Lem}

\begin{Thm}[See \cite{McK25} Thm.\ 7.5]
    \label{thm:isom-dS-AdS}
    Let $(V,\eta)$ be a non-degenerate finite dimensional real quadratic vector space.
    Then the standard action of $\POr(V,\eta)$ on $\dS(V,\eta)$ and $\AdS(V,\eta)$ acts isometrically and we get the following identification with their isometry groups:
    \begin{align}
        \Isom(\dS(V,\eta)) &\cong \POr(V,\eta), &
        \Isom(\AdS(V,\eta)) &\cong \POr(V,\eta).
    \end{align}
\end{Thm}

\begin{Rem}
    \label{rem:isom-dS-AdS-conf-Rd}
    For our pseudo-Euclidean space $\R^{p,q}$ we will use the projective zero quadric $\Mbb^{p,q}$ as its conformal compactification, see later in \Cref{def:conf-comp-metric}.
    For this we need to consider the bigger pseudo-Euclidean vector space $(\R^{p+1,q+1},\eta^{p+1,q+1})$,
    and, the projective space $\Pbb^{d+1} =\Pbb(\R^{p+1,q+1})$.
    Here we get the disjoint decomposition of $\Pbb^{d+1}$ into anti-de Sitter space, conformal compactification and de Sitter space:
    \begin{align}
        \Pbb^{d+1} & = \AdS^{p+1,q} \dcup \Mbb^{p,q} \dcup \dS^{p,q+1}.
    \end{align}
    Note the different signatures on each of the spaces. 
    With \Cref{thm:isom-dS-AdS} we thus get the identifications:
    \begin{align}
        \Isom(\dS^{p,q+1}) &= \POr(p+1,q+1) = \Isom(\AdS^{p+1,q}).
    \end{align}
    In the following, we further want to show that the action of $\POr(p+1,q+1)$ on $\Mbb^{p,q}$ induces conformal diffeomorphisms of $\Mbb^{p,q}$.
    This identification via $\POr(p+1,q+1)$ will thus induce a \emph{correspondence} between (certain) \emph{conformal diffeomorphisms} of the conformal compactification $\Mbb^{p,q}$ of $\R^{p,q}$ and the \emph{isometries} of the corresponding $\AdS^{p+1,q}$-space (or, also the $\dS^{p,q+1}$-space).
\end{Rem}

\subsection{The Conformal Compactification of the Pseudo-Euclidean Space}

\begin{NotLem}[The isometric embedding]
    \label{lem:conf-emb-well-def}
    The following map:
    \begin{align}
        \iota: \R^{p,q} & \to \R^{p+1,q+1}, &  
        \iota(x) &:= \matb{ \frac{1-\eta^{p,q}(x,x)}{2}\\x\\\frac{1+\eta^{p,q}(x,x)}{2}}.
    \end{align}
    is an isometric embedding and satisfies for all $x \in \R^{p,q}$ :
    \begin{align}
        \iota(x) &\neq 0, & \text{ and }  &&\eta^{p+1,q+1}(\iota(x),\iota(x)) & = 0.
    \end{align}
\begin{proof}
    For the latter consider the computation for $x \in \R^{p,q}$:
     \begin{align}
       & 4 \eta^{p+1,q+1}(\iota(x),\iota(x))  \label{eq:conf-emb-well-def}\\
        &= \lp 1-\eta^{p,q}(x,x) \rp^2 + \eta^{p,q}(2x,2x) - \lp 1+\eta^{p,q}(x,x) \rp^2  \\
        &= 1 - 2\eta^{p,q}(x,x) + \eta^{p,q}(x,x)^2 + 4\eta^{p,q}(x,x) - 1 -2\eta^{p,q}(x,x) - \eta^{p,q}(x,x)^2  \\
        &= 0.
    \end{align}
    This shows: $\eta^{p+1,q+1}(\iota(x),\iota(x))  = 0$. $\iota(x) \neq 0$ is clear.
    To see that $\iota$ is an isometric embedding compute the differential (Jacobian matrix):
    \begin{align}
        d\iota_x &= \matb{ -x^\top \Delta^{p,q} \\ I \\ x^\top \Delta^{p,q} }.
    \end{align}
    With this we get:
    \begin{align}
        (d\iota_x)^\top \Delta^{p+1,q+1} d\iota_x 
        & = \matb{ - \Delta^{p,q} x & I &  \Delta^{p,q}x } \Delta^{p+1,q+1} \matb{ -x^\top \Delta^{p,q} \\ I \\ x^\top \Delta^{p,q} } \\
        & = \matb{ - \Delta^{p,q} x & \Delta^{p,q} &  -\Delta^{p,q}x } \matb{ -x^\top \Delta^{p,q} \\ I \\ x^\top \Delta^{p,q} } \\
        & = \Delta^{p,q} x x^\top \Delta^{p,q} + \Delta^{p,q} - \Delta^{p,q} x x^\top \Delta^{p,q} \\
        & = \Delta^{p,q}.
    \end{align}
    This shows that $\iota$ is an isometric embedding.
\end{proof}
\end{NotLem}

\begin{DefsNot}
    For $p,q \ge 0$, $d:=p+q$, we introduce the following notations:
    \begin{enumerate}
        \item We introduce the \emph{affine zero quadric} as the following subspace $Y_0$ of $\R^{p+1,q+1}$ via:
    \begin{align}
        Y_0 &:= \lC y \in \R^{p+1,q+1}\sm\{0\} \st \eta^{p+1,q+1}(y,y) =0 \rC,
    \end{align}
    and endow the tangent spaces for $y \in Y_0$:
    \begin{align}
        \Tan_yY_0 &= \lC v \in \R^{p+1,q+1} \st \eta^{p+1,q+1}(y,v) = 0 \rC \ins \R^{p+1,q+1},
    \end{align}
    with the pull-back metric $\eta^{Y_0}$ of $\eta^{p+1,q+1}$:
    \begin{align}
        \eta^{Y_0}_y(v_1,v_2) &= \eta^{p+1,q+1}(v_1,v_2) \qquad \text{ for } v_1,v_2 \in \Tan_yY_0.
    \end{align}
        \item We also introduce the \emph{double sphere}:
    \begin{align}
        \Sbb^{p,q}:=\Sbb^p \times \Sbb^q &:= \lC y \in \R^{p+1,q+1} \st \sum_{i=0}^p |y^i|^2 = \sum_{j=p+1}^{d+1} |y^j|^2 = 1 \rC \ins Y_0,
    \end{align}
        and endow the tangent spaces for $y \in \Sbb^{p,q}$:
    \begin{align}
        \Tan_y\Sbb^{p,q} 
        &= \lC v \in \R^{p+1,q+1} \st \sum_{i=0}^p y^i \cdot v^i = \sum_{j=p+1}^{d+1} y^j \cdot v^j =0  \rC   \ins   \R^{p+1,q+1}. 
    \end{align}
    with the pull-back metric $\eta^{\Sbb^{p,q}}$ of $\eta^{p+1,q+1}$:
    \begin{align}
        \eta^{\Sbb^{p,q}}_y(v_1,v_2) &= \eta^{p+1,q+1}(v_1,v_2) \qquad \text{ for } v_1,v_2 \in \Tan_y\Sbb^{p,q}.
    \end{align}
\end{enumerate} 
Note that we have the inclusions:
\begin{align}
    \iota(\R^{p,q}) &\ins Y_0 \ins \R^{p+1,q+1}, &
    \Sbb^{p,q} &\ins Y_0 \ins  \R^{p+1,q+1}.
\end{align}
\end{DefsNot}

\begin{NotLem}
    \label{lem:proj-to-Spq}
    Consider the maps:
    \begin{align}
         \rho: Y_0 & \to \R, & \rho(y) &:= \frac{1}{ \sqrt{\sum_{i=0}^p |y^i|^2} }, \\
         \psi: Y_0 & \to \Sbb^{p,q}, & \psi(y) &:= \rho(y) \cdot y.
    \end{align}
    Then $\psi$ is a well-defined map and conformal with conformal factor $\omega_\psi(y)=\rho(y)$.
\begin{proof}
    First note that for $y \in Y_0$ we have: 
    \begin{align}
        0 &=\eta^{p+1,q+1}(y,y)  = \sum_{i=0}^p |y^i|^2 - \sum_{j=p+1}^{d+1} |y^j|^2,
    \end{align}
    and thus:
    \begin{align}
        \rho(y) &= \frac{1}{\sqrt{\sum_{i=0}^p |y^i|^2}} \overset{!}{=} \frac{1}{\sqrt{\sum_{j=p+1}^{d+1} |y^j|^2}}. 
    \end{align}
    This shows that for $y \in Y_0$ we have:
    \begin{align}
        \psi(y) = \rho(y) \cdot y \in \Sbb^p \times \Sbb^q = \Sbb^{p,q}.
    \end{align}
    The differential of $\psi$ at $y \in Y_0$:
    \begin{align}
        d\psi_y: \Tan_yY_0 & \to \Tan_{\psi(y)}\Sbb^{p,q} \ins \Tan_{\psi(y)}\R^{p+1,q+1}.
    \end{align}
    is given with help of the product rule as the matrix:
    \begin{align}
        d\psi_y &= y \cdot \rho'(y) + \rho(y) \cdot I.
    \end{align}
    With this we get for all $y \in Y_0$ and $v_1,v_2 \in \Tan_yY_0$:
    \begin{align}
       & v_1^\top (d\psi_y)^\top \Delta^{p+1,q+1}(d\psi_y) v_2 \\
        &= v_1^\top (y \cdot \rho'(y) + \rho(y) \cdot I)^\top \Delta^{p+1,q+1}(y \cdot \rho'(y) + \rho(y) \cdot I) v_2 \\
        &= \rho(y)^2 \cdot v_1^\top \Delta^{p+1,q+1} v_2 + v_1^ \top\rho'(y)^\top \underbrace{y^\top \Delta^{p+1,q+1} y}_{=0, \text{ as } y \in Y_0} \rho'(y) v_2 \\
        &\qquad   + \rho(y) \cdot v_1^\top \rho'(y)^\top \underbrace{y^\top \Delta^{p+1,q+1}v_2}_{=0, \text{ as } v_2 \in \Tan_yY_0} + \underbrace{v_1^\top\Delta^{p+1,q+1} y}_{=0, \text{ as } v_1 \in \Tan_yY_0} \cdot \rho(y) \cdot \rho'(y)v_2 \\
        &= \rho(y)^2 \cdot v_1^\top \Delta^{p+1,q+1} v_2.
    \end{align}
    This shows the conformal factor of $\rho(y)$.
\end{proof}
\end{NotLem}

\begin{NotLem}
    \label{lem:emb-to-Spq}
    Consider the following map:
    \begin{align}
        \tau: \R^{p,q} &\overset{\iota}{\longrightarrow} Y_0 \overset{\psi}{\longrightarrow} \Sbb^{p,q}, \\
        \tau(x) &:= \psi(\iota(x)) = 
        \frac{1}{\sqrt{ \frac{1}{4} |1-\eta^{p,q}(x,x)|^2 + \sum_{i=1}^p |x^i|^2}} \cdot \matb{ \frac{1-\eta^{p,q}(x,x)}{2}\\x\\\frac{1+\eta^{p,q}(x,x)}{2} }.
    \end{align}
    Then $\tau$ is a well-defined conformal map with conformal factor:
    \begin{align}
        \rho(\iota(x)) &= \frac{1}{\sqrt{\frac{1}{4}|1-\eta^{p,q}(x,x)|^2+\sum_{i=1}^p |x^i|^2}}.
    \end{align}
\end{NotLem}

\begin{Def}[The standard conformal compactification of the pseudo-Euclidean space]
    \label{def:conf-comp-metric}
    For $p,q \ge 0$, $d:=p+q$, the \emph{(standard) conformal compactification} of $(\R^{p,q},\eta^{p,q})$, is defined to be the following projective zero quadric subspace of the $(d+1)$-dimensional (real) projective space:
    \begin{align}
        \Mbb^{p,q} &:= \lC [z]=[z^0:z^1:\dots:z^d:z^{d+1}] \in \Pbb^{d+1} \st \eta^{p+1,q+1}(z,z) = 0 \rC \ins \Pbb^{d+1}.
    \end{align}
    Its metric $\eta^{\Mbb^{p,q}}$ is induced with help of the \emph{double sphere}:
    \begin{align}
        \Sbb^{p,q}:=\Sbb^p \times \Sbb^q &:= \lC y \in \R^{p+1,q+1} \st \sum_{i=0}^p |y^i|^2 = \sum_{j=p+1}^{d+1} |y^j|^2 = 1 \rC \ins \R^{p+1,q+1},
    \end{align}
    and via the $2:1$ locally diffeomorphic surjective map:
    \begin{align}
        \pi: \Sbb^{p,q} &\to \Mbb^{p,q}, & \pm y &\mapsto [y].
    \end{align}
    Note that if $[z] \in \Mbb^{p,q}$ then we can put:
    \begin{align}
        y &:= \psi(z) = \rho(z) \cdot z \in \R^{p+1,q+1},
    \end{align}
    leading to the following properties\footnote{By also multiplying $z$ with $\sgn(z^j)$ for one index $j \in \{0,1, \dots, d, d+1\}$, e.g.\ $j=d+1$, we can, in addition, arrange that $y^j \ge 0$ for this one fixed index $j$. For example, if $q=0$ then $\Sbb^{p,q}=\Sbb^p \times \{\pm 1\}$, and we can always arrange the representative $y$ to have: $y^{d+1} = +1$. In this case: $\Mbb^{p,0} \cong \Sbb^p \times \{+1\}$.}:
    \begin{align}
        y &\in \Sbb^{p,q}, &  [y] &= [z] \in \Mbb^{p,q}.
    \end{align}
    More explicitly, the tangent space of $\Mbb^{p,q}$ at $[z]$ is given by:
    \begin{align}
        \Tan_{[z]}\Mbb^{p,q} &= \lC v \in \R^{p+1,q+1} \st \sum_{i=0}^p z^i \cdot v^i = \sum_{j=p+1}^{d+1} z^j \cdot v^j =0  \rC  \\ 
        &  = \lC v \in \R^{p+1,q+1} \st \sum_{i=0}^p y^i \cdot v^i = \sum_{j=p+1}^{d+1} y^j \cdot v^j =0  \rC  \\
        & = \Tan_y\Sbb^{p,q}  \ins  \Tan_y\R^{p+1,q+1} = \R^{p+1,q+1}. 
    \end{align}
    and the metric of $\Mbb^{p,q}$ at $[z]$ for $v_1,v_2 \in \Tan_{[z]}\Mbb^{p,q}$ by:  
    \begin{align}
        \eta^{\Mbb^{p,q}}_{[z]}(v_1,v_2)  & := \eta^{\Sbb^{p,q}}_y(v_1,v_2) = \eta^{p+1,q+1}(v_1,v_2)= v_1^\top \Delta^{p+1,q+1} v_2 .
    \end{align}
\end{Def}

\begin{Def}[The conformal embedding of the pseudo-Euclidean space into its conformal compactification]
    The \emph{conformal embedding} of $\R^{p,q}$ into $\Mbb^{p,q}$ is defined to be:
    \begin{align}
        [\iota]: \R^{p,q} & \to \Mbb^{p,q}, & [\iota(x)] 
                          &:= \lB \frac{1-\eta^{p,q}(x,x)}{2}:x:\frac{1+\eta^{p,q}(x,x)}{2} \rB \\
                          &&&= [1-\eta^{p,q}(x,x):2x:1+\eta^{p,q}(x,x)].
    \end{align}
    Note that this is a well-defined map, as by \Cref{lem:conf-emb-well-def} we have:
    \begin{align}
        \iota(x) & \neq 0, &  \eta^{p+1,q+1}(\iota(x),\iota(x)) &=0.
    \end{align}
Note that we can also factorize the conformal embedding as follows:
\begin{align}
    [\iota]: \R^{p,q} \overset{\tau}{\longrightarrow} \Sbb^{p,q} \overset{\pi}{\longrightarrow} \Mbb^{p,q}.
\end{align}
leading to the identity:
\begin{align}
    [\tau(x)] &=[\iota(x)],
\end{align}
and showing that the conformal embedding $[\iota]$ is a conformal map with conformal factor:
    \begin{align}
        \omega_{[\iota]}(x) &= \omega_\tau(x) = \rho(\iota(x))= \frac{1}{\sqrt{\frac{1}{4}|1-\eta^{p,q}(x,x)|^2+\sum_{i=1}^p |x^i|^2}}.
    \end{align}
\end{Def}

\subsection{Conformal Transformations of the Conformal Compactification of the Pseudo-Euclidean Space}

\begin{Prp}[See \cite{Scho08} Thm.\ 2.6]
    \label{prp:PO-conf}
Let $p,q \ge 0$. Then every $[\Lambda] \in \POr(p+1,q+1)$ acts as a conformal diffeomorphism on the conformal compactification $\Mbb^{p,q}$ of $\R^{p,q}$ via matrix multiplication and with the conformal factor:
            \begin{align}
                \omega_{[\Lambda]}([z]):=\sqrt{\frac{\sum_{i=0}^p|z^i|^2}{\sum_{i=0}^p|(\Lambda z)^i|^2}}.
            \end{align}
 The inverse of $[\Lambda]$ is given by $[\Lambda^{-1}]$.
 Furthermore, exactly the two matrices $\pm \Lambda \in \Or(p+1,q+1)$ induce the same conformal diffeomorphism on $\Mbb^{p,q}$, thus inducing an embedding/inclusion of groups:
            \begin{align}
                \POr(p+1,q+1) &\ins \ConfDiff(\Mbb^{p,q}).
            \end{align}
\begin{proof}
    For every $\Lambda \in \Or(p+1,q+1)$ it is clear that $\Lambda|_{Y_0}$ is an isometric map from $Y_0$ to $Y_0$. 
    Now define the map:
    \begin{align}
         \psi_\Lambda: \Sbb^{p,q} & \to \Sbb^{p,q}, & \psi_\Lambda(y) &:= \psi(\Lambda y).
    \end{align}
    As the composition:
    \begin{align}
        \psi_\Lambda: \Sbb^{p,q} \ins Y_0 \overset{\Lambda}{\longrightarrow} Y_0 \overset{\psi}{\longrightarrow} \Sbb^{p,q},
    \end{align}
    of the isometry $\Lambda$ (with conformal factor $1$) and the conformal map $\psi$ (with conformal factor $\rho(y)$) also $\psi_\Lambda$ is a conformal map with the conformal factor $\rho(\Lambda y)$.

    If now $[z] \in \Mbb^{p,q}$ then $y:= \rho(z) \cdot z \in \Sbb^{p,q}$. This then shows that:
    \begin{align}
        \psi_{[\Lambda]}: \Mbb^{p,q} & \to \Mbb^{p,q}, & \psi_{[\Lambda]}([z]) &:=[\Lambda z]=[\psi_\Lambda(y)],
    \end{align}
    is a conformal map with conformal factor:
    \begin{align}
        \omega_{[\Lambda]}([z]) &= \rho(\Lambda y) = \rho(\Lambda (\rho(z)z)) 
        = \frac{\rho(\Lambda z)}{\rho(z)}  
        = \sqrt{\frac{\sum_{i=0}^p|z^i|^2}{\sum_{i=0}^p|(\Lambda z)^i|^2}}.
    \end{align}
    Since that action is induced through matrix multiplication it is clear that $[\Lambda^{-1}]$ induces the inverse map to $[\Lambda]$. 

    Now assume that $\Lambda_1$ and $\Lambda_2$ induce the same map on $\Mbb^{p,q}$.
    Then for every $y \in \Sbb^{p,q}$ we have: $[\psi_{\Lambda_1}(y)] = [\psi_{\Lambda_2}(y)]$.
    So $\psi_{\Lambda_1}(y) = \pm \psi_{\Lambda_2}(y)$. 
    Since this holds for all $y \in \Sbb^{p,q}$, we get: $\Lambda_1 = \pm \Lambda_2$.
    This shows the claim.
\end{proof}
\end{Prp}

\begin{Thm}[See \cite{Scho08} Thm.\ 2.6, Thm.\ 2.9, Thm.\ 2.11]
    \label{thm:conf-grp}
    If either $p+q \ge 3$ or $(p,q)=(2,0)$ then the inclusion from \Cref{prp:PO-conf} is already an isomorphism:
            \begin{align}
                \POr(p+1,q+1) &\cong \ConfDiff(\Mbb^{p,q}).  \label{eq:PO-conf-diffeo-iso}
            \end{align}
            This means, in those $(p,q)$-cases, that every conformal diffeomorphism $\varphi: \Mbb^{p,q} \to \Mbb^{p,q}$ 
            is given by the matrix multiplication with an (up to sign) unique matrix $\pm \Lambda \in \Or(p+1,q+1)$, i.e.: $\varphi([z]) = [\Lambda z]$ for all $[z] \in \Mbb^{p,q}$.
\end{Thm}

\begin{Thm}[See \cite{Scho08} Thm.\ 2.6, Thm.\ 2.9, Thm.\ 2.11]
    \label{thm:conf-grp-ext}
    In addition to the isomorphism in \Cref{eq:PO-conf-diffeo-iso} in \Cref{thm:conf-grp} we get the following stronger statements:
    \begin{enumerate}
        \item Let $p+q \ge 3$, then for every conformal map $\varphi:U \to \R^{p,q}$, defined on any connected open subset $U \ins \R^{p,q}$, there exists an (up to sign) unique matrix $\pm\Lambda \in \Or(p+1,q+1)$ such that the following diagram commutes:
            \begin{align}
                \xymatrix{
                    U \ar_{[\iota]}[d] \ar^{\varphi}[r] & \R^{p,q} \ar^{[\iota]}[d]  \\
                \Mbb^{p,q} \ar_{[\Lambda]}[r]  & \Mbb^{p,q}.
            }  \label{eq:conf-comm-diag}
            \end{align}
            i.e.\ for all $x \in U$ we have:
            \begin{align}
                [\iota\circ\varphi(x)] &= [\Lambda \iota(x)].
            \end{align}
            In particular, $\varphi$ is injective.
        \item  Let $(p,q)=(2,0)$, then for every \emph{injective} conformal map $\varphi:U \to \R^{2,0}$, defined either on $U=\R^{2,0}$ or on any \emph{punctured plane} $U =\R^{2,0} \sm \{\tilde x\}$, there exists an (up to sign) unique matrix $\pm\Lambda \in \Or(3,1)$ such that the corresponding diagram from \ref{eq:conf-comm-diag} commutes.\footnote{Note that, for the case $(p,q)=(2,0)$, the injectivity of $\varphi$ needs to be assumed and we also can only allow for open subsets $U$ of $\R^{2,0}$ where at most one point is removed from $\R^{2,0}$.}
    \end{enumerate}
\end{Thm}

\begin{Rem}
For $(p,q)=(1,1)$ the group $\ConfDiff(\Mbb^{p,q})$ is much bigger than $\POr(p+1,q+1)$. 
For details see \cite{Scho08} section 2.5.
\end{Rem}

\begin{Eg}[The inversion at the pseudo-sphere]
    \label{eg:inversion-2}
    We continue the discussion about the \emph{inversion (at the pseudo-sphere)} $\invsp$ on $\R^{p,q}$ from \Cref{eg:inversion}: 
    \begin{align}
        \invsp=\invsp^{p,q}: U &\to \R^{p,q}, & \invsp^{p,q}(x) &:= \frac{x}{\eta^{p,q}(x,x)}.
    \end{align}
    defined on $U:=\lC x \in \R^{p,q} \st \eta^{p,q}(x,x) \neq 0 \rC$.
    Consider the $(d+2) \times (d+2)$-matrix: 
    \begin{align}
        \Lambda_{\invsp} &:= \matb{ -1 & 0 & 0 \\ 0 & I & 0 \\ 0 & 0 & 1 }.
    \end{align}
    We now claim that the map:
    \begin{align}
        \bar \invsp:= [\Lambda_{\invsp}]: \Mbb^{p,q} &\to \Mbb^{p,q}, & [\Lambda_{\invsp}][z] &:= [\Lambda_{\invsp}z],
    \end{align}
    is the conformal extension of $\invsp$ from $U$ to whole $\Mbb^{p,q}$.
    Indeed, first note that $\Lambda_{\invsp} \in \Or(p+1,q+1)$ and $\det \Lambda_{\invsp} = -1$ and $\det (- \Lambda_\invsp) =(-1)^{d+1}$.
    Then compute:
    \begin{align}
      \iota(\invsp(x)) 
      &= \frac{1}{2}\matb{ 1-\eta(\invsp(x),\invsp(x))\\ 2 \invsp(x) \\ 1+\eta(\invsp(x),\invsp(x))} \\
      &= \frac{1}{2}\matb{ 1 - \frac{1}{\eta(x,x)} \\ \frac{2x}{\eta(x,x)} \\ 1 + \frac{1}{\eta(x,x)}} \\
      &= \frac{1}{2}\frac{1}{\eta(x,x)} \matb{ \eta(x,x) -1 \\ 2x \\ \eta(x,x) + 1 } \\
      &= \frac{1}{2}\frac{1}{\eta(x,x)} \matb{ -1 & 0 & 0 \\ 0 & I & 0 \\ 0 & 0 & 1 } \matb{ 1-\eta(x,x) \\ 2x \\ 1+\eta(x,x) } \\
      &= \frac{1}{\eta(x,x)} \cdot \Lambda_{\invsp}\iota(x),
    \end{align}
    which shows the claim:
    \begin{align}
        [\Lambda_{\invsp}\iota(x)] & = [\iota(\invsp(x))] \in \Mbb^{p,q}.
    \end{align}
    Note that for $x \in \R^{p,q}$ with $\eta(x,x) = 0 $ we get: 
    \begin{align} 
        \iota(x) &=[1:2x:1], &  \bar\invsp(\iota(x)) &=[-1:2x:1],  \\
        \iota(0) &= [1:0:1], & \bar\invsp(\iota(0)) &=[-1:0:1].
    \end{align}
\end{Eg}

\begin{Eg}[Linear conformal transformations, see \cite{Scho08} Thm.\ 2.9]
    \label{eg:lin-conf-diff-2}
    We continue from \Cref{eg:aff-conf-diff} and \Cref{thm:aff-conf-diff} for the linear conformal map: 
    $x \mapsto Ax$ with the matrix $A=c\Lambda$, where $A=c\Lambda \in \COr(p,q)$ with $c >0 $ and $\Lambda \in \Or(p,q)$. Then we can define the $(d+2)\times(d+2)$-matrix:
    \begin{align}
        \Gamma_{c\Lambda} & := \matb{ \frac{1+c^2}{2c} & 0 & \frac{1-c^2}{2c} \\
                            0 & \Lambda & 0 \\
                        \frac{1-c^2}{2c} & 0 & \frac{1+c^2}{2c}}.
    \end{align}
    Then $\Gamma_{c\Lambda} \in \Or(p+1,q+1)$ as:
    \begin{align}
        \Gamma_{c\Lambda}^\top \Delta^{p+1,q+1} \Gamma_{c\Lambda} 
        &= \matb{\frac{1+c^2}{2c}&0&\frac{1-c^2}{2c}\\0&\Lambda^\top&0\\\frac{1-c^2}{2c}&0&\frac{1+c^2}{2c}}  \matb{1&0&0\\0&\Delta^{p,q}&0\\0&0&-1} \matb{\frac{1+c^2}{2c}&0&\frac{1-c^2}{2c}\\0&\Lambda&0\\\frac{1-c^2}{2c}&0&\frac{1+c^2}{2c}} \\
        &= \matb{\frac{1+c^2}{2c}&0&\frac{1-c^2}{2c}\\0&\Lambda^\top&0\\\frac{1-c^2}{2c}&0&\frac{1+c^2}{2c}} \matb{\frac{1+c^2}{2c}&0&\frac{1-c^2}{2c}\\0&\Delta^{p,q}\Lambda&0\\-\frac{1-c^2}{2c}&0&-\frac{1+c^2}{2c}} \\
        &= \matb{\lp\frac{1+c^2}{2c}\rp^2-\lp\frac{1-c^2}{2c}\rp^2 &0&0\\0&\Lambda^\top\Delta^{p,q}\Lambda&0\\0&0&\lp\frac{1-c^2}{2c}\rp^2-\lp\frac{1+c^2}{2c}\rp^2  } \\
        &= \matb{1&0&0\\0&\Delta^{p,q}&0\\0&0&-1} \\
        &= \Delta^{p+1,q+1}.
    \end{align}
    Also note that if $\Lambda \in \Or^0(p,q)$ then $\Gamma_{c\Lambda} \in \Or^0(p+1,q+1)$ as $c \to 1$ provides a path to the identity component.\footnote{The reverse is also true: $\Gamma_{c\Lambda} \in \Or^0(p+1,q+1) \implies \Lambda \in \Or^0(p,q)$. } 
    We now claim that:
    \begin{align}
        [\Gamma_{c\Lambda}]: \Mbb^{p,q} &\to \Mbb^{p,q}, & [\Gamma_{c\Lambda}][z] &:= [\Gamma_{c\Lambda}z],
    \end{align}
    is the conformal extension of $c\Lambda$ from $\R^{p,q}$ to $\Mbb^{p,q}$.
    \begin{align}
        \Gamma_{c\Lambda}\iota(x) & = \matb{\frac{1+c^2}{2c}&0&\frac{1-c^2}{2c}\\0&\Lambda&0\\\frac{1-c^2}{2c}&0&\frac{1+c^2}{2c}} \matb{\frac{1-\eta^{p,q}(x,x)}{2}\\x\\\frac{1+\eta^{p,q}(x,x)}{2}}  \\
        &= \matb{ \lp\frac{1+c^2}{2c}\rp\lp\frac{1-\eta^{p,q}(x,x)}{2}\rp+\lp\frac{1-c^2}{2c}\rp\lp\frac{1+\eta^{p,q}(x,x)}{2}\rp  \\ \Lambda x \\ \lp\frac{1-c^2}{2c}\rp\lp\frac{1-\eta^{p,q}(x,x)}{2}\rp+\lp\frac{1+c^2}{2c}\rp\lp\frac{1+\eta^{p,q}(x,x)}{2}\rp } \\
        &= \frac{1}{4c}\matb{ 1+c^2 - \eta^{p,q}(x,x) - c^2 \cdot \eta^{p,q}(x,x) + 1 - c^2 +  \eta^{p,q}(x,x) - c^2 \cdot \eta^{p,q}(x,x)  \\ 4c\Lambda x \\ 1-c^2 - \eta^{p,q}(x,x) + c^2 \cdot \eta^{p,q}(x,x) + 1 + c^2 +  \eta^{p,q}(x,x) + c^2 \cdot \eta^{p,q}(x,x) } \\
        &= \frac{1}{4c}\matb{ 2  - 2 \cdot c^2 \cdot \eta^{p,q}(x,x)  \\ 4c\Lambda x \\ 2  + 2 \cdot c^2 \cdot \eta^{p,q}(x,x) } \\
        &= \frac{1}{c} \matb{ \frac{1-\eta^{p,q}(c\Lambda x,c\Lambda x)}{2}\\ c\Lambda x\\ \frac{1+\eta^{p,q}(c\Lambda x,c\Lambda x)}{2}} \\
        &= \frac{1}{c}\cdot \iota\lp c\Lambda x \rp.
    \end{align}
    This shows:
    \begin{align}
        [\Gamma_{c\Lambda}\iota(x)] & = [\iota(c\Lambda x)] \in \Mbb^{p,q},
    \end{align}
    and thus the claim.
\end{Eg}

\begin{Eg}[Translations, see \cite{Scho08} Thm.\ 2.9]
    \label{eg:trans-conf-diff-2}
    We now continue from \Cref{eg:aff-conf-diff} and \Cref{thm:aff-conf-diff} for case of translation: 
    $x \mapsto x + b$ for $b \in \R^{p,q}$.
    Then we can define the $(d+2)\times(d+2)$-matrix:
    \begin{align}
        \Gamma_{I,b} &:= I + \Omega_b \\
        &:= I + \matb{-\frac{1}{2}\eta^{p,q}(b,b)&-b^\top\Delta^{p,q}&-\frac{1}{2}\eta^{p,q}(b,b)\\b&0&b\\\frac{1}{2}\eta^{p,q}(b,b)&b^\top\Delta^{p,q}&\frac{1}{2}\eta^{p,q}(b,b)} \\
        &= \matb{1-\frac{1}{2}\eta^{p,q}(b,b)&-b^\top\Delta^{p,q}&-\frac{1}{2}\eta^{p,q}(b,b)\\b&I&b\\\frac{1}{2}\eta^{p,q}(b,b)&b^\top\Delta^{p,q}&1+\frac{1}{2}\eta^{p,q}(b,b)}.
    \end{align}
    We first show that $\Gamma_{I,b} \in \Or(p+1,q+1)$. 
    For this consider:
    \begin{align}
    \Gamma_{I,b}^\top \Delta^{p+1,q+1} \Gamma_{I,b} 
    &= \Delta^{p+1,q+1} + \Omega_b^\top\Delta^{p+1,q+1}+ \Delta^{p+1,q+1}\Omega_b + \Omega_b^\top \Delta^{p+1,q+1} \Omega_b.
    \end{align}
    We compute terms separately:
    \begin{align}
        \Delta^{p+1,q+1}\Omega_b 
        &= \matb{1&0&0\\0&\Delta^{p,q}&0\\0&0&-1}\matb{-\frac{1}{2}\eta^{p,q}(b,b)&-b^\top\Delta^{p,q}&-\frac{1}{2}\eta^{p,q}(b,b)\\b&0&b\\\frac{1}{2}\eta^{p,q}(b,b)&b^\top\Delta^{p,q}&\frac{1}{2}\eta^{p,q}(b,b)} \\
        &= \matb{-\frac{1}{2}\eta^{p,q}(b,b)&-b^\top\Delta^{p,q}&-\frac{1}{2}\eta^{p,q}(b,b)\\\Delta^{p,q}b&0&\Delta^{p,q}b\\-\frac{1}{2}\eta^{p,q}(b,b)&-b^\top\Delta^{p,q}&-\frac{1}{2}\eta^{p,q}(b,b)}, \\
     \Omega_b^\top \Delta^{p+1,q+1} &= \lp \Delta^{p+1,q+1}\Omega_b \rp^\top \\
        &= \matb{-\frac{1}{2}\eta^{p,q}(b,b)& b^\top\Delta^{p,q}&-\frac{1}{2}\eta^{p,q}(b,b)\\-\Delta^{p,q}b&0&-\Delta^{p,q}b\\-\frac{1}{2}\eta^{p,q}(b,b)&b^\top\Delta^{p,q}&-\frac{1}{2}\eta^{p,q}(b,b)}.
    \end{align}
 With this we compute:
    \begin{align}
        \Omega_b^\top \Delta^{p+1,q+1} + \Delta^{p+1,q+1}\Omega_b &= - \eta^{p,q}(b,b) \cdot \matb{1&0&1\\0&0&0\\1&0&1},\\
    \end{align}
 We also get:
    \begin{align}
  \Omega_b^\top \Delta^{p+1,q+1} \Omega_b 
  &= \matb{-\frac{1}{2}\eta^{p,q}(b,b)& b^\top\Delta^{p,q}&-\frac{1}{2}\eta^{p,q}(b,b)\\-\Delta^{p,q}b&0&-\Delta^{p,q}b\\-\frac{1}{2}\eta^{p,q}(b,b)&b^\top\Delta^{p,q}&-\frac{1}{2}\eta^{p,q}(b,b)} \matb{-\frac{1}{2}\eta^{p,q}(b,b)&-b^\top\Delta^{p,q}&-\frac{1}{2}\eta^{p,q}(b,b)\\b&0&b\\\frac{1}{2}\eta^{p,q}(b,b)&b^\top\Delta^{p,q}&\frac{1}{2}\eta^{p,q}(b,b)} \\
  &= \matb{b^\top\Delta^{p,q}b&0&b^\top\Delta^{p,q}b\\0&0&0\\b^\top\Delta^{p,q}b&0&b^\top\Delta^{p,q}b} \\
  &= \eta^{p,q}(b,b) \cdot \matb{1&0&1\\0&0&0\\1&0&1}.
    \end{align}
  Together this shows:
  \begin{align}
    \Gamma_{I,b}^\top \Delta^{p+1,q+1} \Gamma_{I,b} 
    &= \Delta^{p+1,q+1} + \Omega_b^\top\Delta^{p+1,q+1}+ \Delta^{p+1,q+1}\Omega_b + \Omega_b^\top \Delta^{p+1,q+1} \Omega_b \\
    &= \Delta^{p+1,q+1} - \eta^{p,q}(b,b) \cdot \matb{1&0&1\\0&0&0\\1&0&1} + \eta^{p,q}(b,b) \cdot \matb{1&0&1\\0&0&0\\1&0&1} \\
    &= \Delta^{p+1,q+1}.
  \end{align}
This thus shows the claim: $\Gamma_{I,b} \in \Or(p+1,q+1)$.
Furthermore, for $b \to 0$ we see that $\Gamma_{I,b} \to I$. 
This thus even shows that: $\Gamma_{I,b} \in \Or^0(p+1,q+1)$.

We now claim that $[\Gamma_{I,b}]$ is the conformal extension of the translation map $x \mapsto x+b$ from $\R^{p,q}$ to $\Mbb^{p,q}$.
For this compute:
\begin{align}
    \Gamma_{I,b}\iota(x) &= \lp I+\Omega_b \rp \iota(x) \\
    &=\iota(x) + \Omega_b \iota(x),
\end{align}
    with:
\begin{align}
    \Omega_b  \iota(x)
    &= \frac{1}{2} \matb{-\frac{1}{2}\eta^{p,q}(b,b)&-b^\top\Delta^{p,q}&-\frac{1}{2}\eta^{p,q}(b,b)\\b&0&b\\\frac{1}{2}\eta^{p,q}(b,b)&b^\top\Delta^{p,q}&\frac{1}{2}\eta^{p,q}(b,b)} \matb{1-\eta^{p,q}(x,x)\\2x\\1+\eta^{p,q}(x,x)} \\
    &= \frac{1}{2}\matb{-\eta^{p,q}(b,b)-2b^\top\Delta^{p,q}x\\2b\\\eta^{p,q}(b,b)+2b^\top\Delta^{p,q}x} \\
    &= \frac{1}{2}\matb{-\eta^{p,q}(b,b)-2\eta^{p,q}(b,x)\\2b\\\eta^{p,q}(b,b)+2\eta^{p,q}(b,x)}.
\end{align}
With this we get:
\begin{align}
    \iota(x+b) 
    &= \frac{1}{2}\matb{1-\eta^{p,q}(x+b,x+b)\\2(x+b)\\1+\eta^{p,q}(x+b,x+b)} \\
    &= \frac{1}{2}\matb{1-\eta^{p,q}(x,x)-\eta^{p,q}(b,b)-2\eta^{p,q}(b,x)\\2x+2b\\1+\eta^{p,q}(x,x)+\eta^{p,q}(b,b)+2\eta^{p,q}(b,x)} \\
    &= \frac{1}{2}\matb{1-\eta^{p,q}(x,x)\\2x\\1+\eta^{p,q}(x,x)} + \frac{1}{2}\matb{-\eta^{p,q}(b,b)-2\eta^{p,q}(b,x)\\2b\\\eta^{p,q}(b,b)+2\eta^{p,q}(b,x)} \\
    &= \iota(x) + \Omega_b\iota(x) \\
    &= \Gamma_{I,b}\iota(x).
\end{align}
This then implies:
\begin{align}
    [\Gamma_{I,b}\iota(x)] &= [\iota(x+b)] \in \Mbb^{p,q}.
\end{align}
This shows that $[\Gamma_{I,b}]$ conformally extends the translation map $x \mapsto x+b$ from $\R^{p,q}$ to $\Mbb^{p,q}$.
\end{Eg}

\begin{Eg}[Affine conformal transformations, see \cite{Scho08} Thm.\ 2.9]
    \label{eg:aff-conf-diff-2}
We continue from \Cref{eg:aff-conf-diff} and \Cref{thm:aff-conf-diff} for the affine conformal map: 
$x \mapsto Ax+b$ with the matrix $A=c\Lambda$ and $b \in \R^{p,q}$, where $A=c\Lambda \in \COr(p,q)$ with $c >0 $ and $\Lambda \in \Or(p,q)$. Then we can define the $(d+2)\times(d+2)$-matrix:
    \begin{align}
        \Gamma_{c\Lambda,b} &:= \Gamma_{I,b}\Gamma_{c\Lambda} \\
        &= \matb{1-\frac{1}{2}\eta^{p,q}(b,b)&-b^\top\Delta^{p,q}&-\frac{1}{2}\eta^{p,q}(b,b)\\b&I&b\\\frac{1}{2}\eta^{p,q}(b,b)&b^\top\Delta^{p,q}&1+\frac{1}{2}\eta^{p,q}(b,b)} \matb{\frac{1+c^2}{2c}&0&\frac{1-c^2}{2c}\\0&\Lambda&0\\\frac{1-c^2}{2c}&0&\frac{1+c^2}{2c}}.
    \end{align}
    Then we have: $\Gamma_{c\Lambda} \in \Or(p+1,q+1)$. This is clear, as both matrices $\Gamma_{I,b},\Gamma_{c\Lambda} \in \Or(p+1,q+1)$ by \Cref{eg:lin-conf-diff-2} and \Cref{eg:trans-conf-diff-2}.
    It thus also clear that the matrix 
    $[\Gamma_{c\Lambda,b}]$ extends the affine map $x \mapsto c\Lambda x +b $ from $\R^{p,q}$ to $\Mbb^{p,q}$.
\end{Eg}

\begin{Eg}[Special conformal transformations, see \cite{Scho08} Thm.\ 2.9]
    \label{eg:spec-conf-trafo-2}
    We continue from \Cref{eg:spec-conf-trafo} about the \emph{special conformal transformations}.
    Recall that we defined the special conformal transformations for $b \in \R^{p,q}$ on an open subset $U_b \ins \R^{p,q}$ as:
    \begin{align}
        \sigma_b: U_b &\to \R^{p,q}, & 
        \sigma_b(x) &:=  \frac{x - \eta(x,x)\cdot b}{1-2 \cdot \eta(x,b) + \eta(b,b)\cdot\eta(x,x)} = \invsp\lp \invsp(x) - b \rp,
    \end{align}
    where $\invsp$ is the inversion at the pseudo-sphere from \Cref{eg:inversion} and \Cref{eg:inversion-2}. 
    With the latter representation and \Cref{eg:trans-conf-diff-2} we immediately get that the $(d+2)\times(d+2)$-matrix:
    \begin{align}
        \Sigma_b &:= \Lambda_\invsp \Gamma_{I,-b} \Lambda_\invsp \\
                 &=  \matb{-1&0&0\\0&I&0\\0&0&1} \matb{1-\frac{1}{2}\eta^{p,q}(b,b)&b^\top\Delta^{p,q}&-\frac{1}{2}\eta^{p,q}(b,b)\\-b&I&-b\\\frac{1}{2}\eta^{p,q}(b,b)&-b^\top\Delta^{p,q}&1+\frac{1}{2}\eta^{p,q}(b,b)} \matb{-1&0&0\\0&I&0\\0&0&1} \\
                 &= \matb{1-\frac{1}{2}\eta^{p,q}(b,b)&-b^\top\Delta^{p,q}&\frac{1}{2}\eta^{p,q}(b,b)\\b&I&-b\\-\frac{1}{2}\eta^{p,q}(b,b)&-b^\top\Delta^{p,q}&1+\frac{1}{2}\eta^{p,q}(b,b)}
    \end{align}
    lies in $\Or(p+1,q+1)$ and that $[\Sigma_b]$ extends the special conformal transformation $\sigma_b$ from $\R^{p,q}$ to $\Mbb^{p,q}$. 
    Furthermore, for $b \to 0$ we see that $\Sigma_b \to I$. This then even shows that: $\Sigma_b \in \Or^0(p+1,q+1)$.
\end{Eg}

\subsection{The Definition of the Conformal Group of the Pseudo-Euclidean Space}

\begin{Def}[The global and restricted conformal group of the pseudo-Euclidean space]
    Based on \Cref{thm:conf-grp} we define for all $p,q \ge 0$ the 
    \emph{global conformal group} of $\R^{p,q}$ to be:
    \begin{align}
        \Conf_g(\R^{p,q}) &:=\POr(p+1,q+1),
    \end{align}
    and the \emph{(restricted) conformal group} $\Conf(\R^{p,q})$ of $\R^{p,q}$ to be the connected component of the identity of $\POr(p+1,q+1)$:
    \begin{align}
        \Conf(\R^{p,q}) &:= \Conf_r(\R^{p,q}) := \Conf_g^0(\R^{p,q}) =\POr^0(p+1,q+1).
    \end{align}
    Note that by \Cref{prp:PO-conf} we always have the inclusions:
    \begin{align}
        \Conf_r(\R^{p,q}) \ins \Conf_g(\R^{p,q}) \ins \ConfDiff(\Mbb^{p,q}).
    \end{align}
\end{Def}

\begin{Thm}
    $\Conf(\R^{p,q})$ contains all affine conformal transformations $x \mapsto Ax+b$ of $\R^{p,q}$ where $A =c\Lambda \in \COr^0(p,q)$ with $c > 0$ and $\Lambda \in \Or^0(p,q)$ and $b \in \R^{p,q}$ in form of the map $[\Gamma_{c\Lambda,b}]:\Mbb^{p,q}\to\Mbb^{p,q}$ from \Cref{eg:aff-conf-diff-2}, and, also all the special conformal transformations $\sigma_b$ in form of the map $[\Sigma_b]:\Mbb^{p,q}\to\Mbb^{p,q}$ from \Cref{eg:spec-conf-trafo-2}.
\end{Thm}

\subsection{A (Partial) Parameterization of the Anti-de Sitter Space}
\label{sec:AdS-param}

\begin{Prp}
    \label{prp:para-ads}
        Let $p,q \ge 0$ and $d:=p+q$. 
Consider the following space:
\begin{align}
    \Abb^{p+1,q} := \R_{> 0} \times \R^{p,q},
\end{align}
which we endow with the metric $\eta^{\Abb^{p+1,q}}$, which is given at $x \in \Abb^{p+1,q}$ and with $v_1,v_2 \in \Tan_x \Abb^{p+1,q} = \R^{p+1,q}$ as follows:
    \begin{align}
        \eta^{\Abb^{p+1,q}}_x(v_1,v_2) &:=   \frac{\eta^{p+1,q}(v_1, v_2)}{|x^0|^2}.
    \end{align}
Then consider the map:
    \begin{align}
        \phi : \Abb^{p+1,q} &\to  \Rbb^{p+1,q+1}, \quad x \mapsto \phi(x) =: y, \\
\\
         \phi^0(x) &:= \frac{1 - \eta^{p+1,q}(x,x)}{2x^0} 
            = \frac{1}{2x^0} - \frac{x^0}{2} -  \frac{1}{2x^0}\eta^{p,q}(x^{1:d},x^{1:d}),\\               
          \phi^{1:d}(x) &:= \frac{x^{1:d}}{x^0}, \\
         \phi^{d+1}(x) &:= \frac{1 + \eta^{p+1,q}(x,x)}{2x^0} 
        = \frac{1}{2x^0} + \frac{x^0}{2} +  \frac{1}{2x^0}\eta^{p,q}(x^{1:d},x^{1:d}).
    \end{align}
Then $\phi$ is a well-defined isometric embedding and its range coincides with the subspace:
\begin{align}
    \tAdS^{p+1,q} &:= \lC y \in \R^{p+1,q+1} \st \eta^{p+1,q+1}(y,y) = -1 \;\land\; y^0+y^{d+1} > 0 \rC.
\end{align}
An inverse $\phi^{-1}:\tAdS^{p+1,q} \to \Abb^{p+1,q}$ of $\phi$ is given by:
    \begin{align}
        x^0:= (\phi^{-1})^0(y) &= \frac{1}{y^0 + y^{d+1}} > 0, \\
        x^{1:d} := (\phi^{-1})^{1:d}(y) &= \frac{y^{1:d}}{y^0 + y^{d+1}}.
    \end{align}
\begin{proof}
    We first show that $\phi$ is well-defined. For this we compute for $y:=\phi(x)$:
\begin{align}
    4|x^0|^2 \cdot \eta^{p+1,q+1}(y,y) 
    &= \lp 1 - \eta^{p+1,q}(x,x) \rp^2 + 4 \eta^{p,q}(x^{1:d},x^{1:d}) - \lp 1 + \eta^{p+1,q}(x,x) \rp^2 \\
    &= -4 \eta^{p+1,q}(x,x) + 4 \eta^{p,q}(x^{1:d},x^{1:d}) \\
    &= -4 |x^0|^2,
\end{align}
implying: $\eta^{p+1,q+1}(y,y) = -1$. Furthermore:
\begin{align}
    y^0+y^{d+1} &= \frac{1 - \eta^{p+1,q}(x,x)}{2x^0} + \frac{1 + \eta^{p+1,q}(x,x)}{2x^0} =  \frac{1}{x^0} > 0.
\end{align}
Together, this shows: $\phi(x) = y \in \tAdS^{p+1,q} \ins \R^{p+1,q+1}$.
It is easy to see that $\phi^{-1}$ is the left-inverse to $\phi$, i.e.:
\begin{align}
    \phi^{-1} \circ \phi(x) &= x \in \Abb^{p+1,q}.
\end{align}
To show that $\phi^{-1}$ is also a right-invers let $y \in \tAdS^{p+1,q}$ and $x:=\phi^{-1}(y)$. 
Then compute:
\begin{align}
 \phi^{1:d}(x) &= \frac{1}{x^0} x^{1:d} = \lp y^0+y^{d+1} \rp \cdot \frac{y^{1:d}}{y^0+y^{d+1} } = y^{1:d}, \\
 \phi^0(x) &= \frac{1}{2x^0} - \frac{x^0}{2} -  \frac{1}{2x^0}\eta^{p,q}(x^{1:d},x^{1:d}) \\
     &= \frac{y^0+y^{d+1}}{2} - \frac{1}{2\lp y^0+y^{d+1}\rp} -  \frac{1}{2\lp y^0+y^{d+1}\rp}\eta^{p,q}(y^{1:d},y^{1:d}) \\
    &= \frac{1}{2}\lp \lp y^0 + y^{d+1} \rp - \frac{1}{y^0+y^{d+1}}\lp 1+ \eta^{p,q}(y^{1:d},y^{1:d}) \rp \rp \\
    &= \frac{1}{2}\lp \lp y^0 + y^{d+1} \rp - \frac{1}{y^0+y^{d+1}}\lp |y^{d+1}|^2-|y^0|^2 \rp \rp \\
    &= \frac{1}{2}\lp \lp y^0 + y^{d+1} \rp - \lp y^{d+1}-y^0 \rp \rp \\
    &= y^0.
\end{align}
Similarly:
\begin{align}
    \phi^{d+1}(x) &= \frac{1}{2x^0} + \frac{x^0}{2} +  \frac{1}{2x^0}\eta^{p,q}(x^{1:d},x^{1:d}) \\
     &= \frac{y^0+y^{d+1}}{2} + \frac{1}{2\lp y^0+y^{d+1}\rp} +  \frac{1}{2\lp y^0+y^{d+1}\rp}\eta^{p,q}(y^{1:d},y^{1:d}) \\
    &= \frac{1}{2}\lp \lp y^0 + y^{d+1} \rp + \frac{1}{y^0+y^{d+1}}\lp 1+ \eta^{p,q}(y^{1:d},y^{1:d}) \rp \rp \\
    &= \frac{1}{2}\lp \lp y^0 + y^{d+1} \rp + \frac{1}{y^0+y^{d+1}}\lp |y^{d+1}|^2-|y^0|^2 \rp \rp \\
    &= \frac{1}{2}\lp \lp y^0 + y^{d+1} \rp + \lp y^{d+1}-y^0 \rp \rp \\
    &= y^{d+1}.
\end{align}
This shows that:
\begin{align}
    \phi \circ \phi^{-1}(y) &= y \in \tAdS^{p+1,q}.
\end{align}

We now compute the differential of $\phi$:
    \begin{align}
        d\phi_x &= \matb{ -\frac{1}{2|x^0|^2}-\frac{1}{2}+\frac{\eta^{p,q}(x^{1:d},x^{1:d})}{2|x^0|^2} & -\frac{1}{x^0}(x^{1:d})^\top \Delta^{p,q} \\ -\frac{x^{1:d}}{|x^0|^2} & \frac{I_d}{x^0}\\ -\frac{1}{2|x^0|^2}+\frac{1}{2}-\frac{\eta^{p,q}(x^{1:d},x^{1:d})}{2|x^0|^2} & \frac{1}{x^0}(x^{1:d})^\top \Delta^{p,q} }
    \end{align}
and the metric at $x \in \Abb^{p+1,q}$:
    \begin{align}
         (d\phi_x)^\top \Delta^{p+1,q+1} d\phi_x = (d\phi_x)^\top \matb{1&0&0\\0&\Delta^{p,q}&0\\0&0&-1} d\phi_x.
    \end{align}
    We compute the entries of the last matrix separately. First, top left entry:
    \begin{align}
     &   \frac{1}{4}\lp1-\frac{\eta^{p,q}(x^{1:d},x^{1:d})}{|x^0|^2}+\frac{1}{|x^0|^2}\rp^2 + \frac{\eta^{p,q}(x^{1:d},x^{1:d})}{|x^0|^4} - \frac{1}{4}\lp1-\frac{\eta^{p,q}(x^{1:d},x^{1:d})}{|x^0|^2}-\frac{1}{|x^0|^2}\rp^2 \\
     &= \frac{1}{|x^0|^2}\lp1-\frac{\eta^{p,q}(x^{1:d},x^{1:d})}{|x^0|^2}\rp + \frac{\eta^{p,q}(x^{1:d},x^{1:d})}{|x^0|^4}  \\
     &= \frac{1}{|x^0|^2}.
    \end{align}
    Note that in the middle we used the formula: $(a+b)^2-(a-b)^2 = 4ab$.
    Next, top right entry:
    \begin{align}
        & \frac{1}{2}\lp\lp1-\frac{\eta^{p,q}(x^{1:d},x^{1:d})}{|x^0|^2}\rp+\frac{1}{|x^0|^2}\rp \frac{(x^{1:p})^\top\Delta^{p,q}}{x^0} -\frac{(x^{1:d})^\top\Delta^{p,q}}{|x^0|^3} \\
        &- \frac{1}{2}\lp\lp1-\frac{\eta^{p,q}(x^{1:d},x^{1:d})}{|x^0|^2}\rp-\frac{1}{|x^0|^2}\rp \frac{(x^{1:d})^\top\Delta^{p,q}}{x^0} \\
        &= \frac{1}{2}\lp \frac{2}{|x^0|^2} \rp \frac{(x^{1:d})^\top\Delta^{p,q}}{x^0} -\frac{(x^{1:d})^\top\Delta^{p,q}}{|x^0|^3} \\
        &= 0.
    \end{align}
    By symmetry also the bottom left equals $0$.
    Finally, bottom right entry:
    \begin{align}
        \frac{\Delta^{p,q}x^{1:d}}{x^0}\frac{(x^{1:d})^\top\Delta^{p,q}}{x^0} + \frac{I_p}{x^0}\Delta^{p,q}\frac{I_p}{x^0} - \frac{\Delta^{p,q}x^{1:d}}{x^0}\frac{(x^{1:d})^\top\Delta^{p,q}}{x^0} & = \frac{\Delta^{p,q}}{|x^0|^2}.
    \end{align}
    Together we get:
    \begin{align}
        (d\phi_x)^\top \Delta^{p+1,q+1} d\phi_x &= \frac{1}{|x^0|^2} \matb{1&0\\0&\Delta^{p,q}} = \frac{1}{|x^0|^2} \Delta^{p+1,q}.
    \end{align}
    This shows the claim.   
\end{proof}
\end{Prp}

\begin{Thm}
    \label{thm:para-ads}
    The map:
    \begin{align} 
        \Abb^{p+1,q} \cong \tAdS^{p+1,q} &\to \AdS^{p+1,q}, & x &\mapsto \phi(x)=:y \mapsto [y],
    \end{align}
    is an isometric embedding. The complement of its range is given by the closed subspace:
    \begin{align}
       \Zcal:= \lC [z] \in \AdS^{p+1,q} \st z^0 + z^{d+1} = 0 \rC.
    \end{align}
    The left-inverse for $[z] \in \AdS^{p+1,q}$ (with $z^0 + z^{d+1} \neq 0$) is given by:
    \begin{align}
        y &:= \frac{\sgn(z^0+z^{d+1})}{\sqrt{|\eta^{p+1,q+1}(z,z)|}} \cdot z \in \tAdS^{p+1,q}, 
    \end{align}
    and to get $x \in \Abb^{p+1,q}$ we put:
    \begin{align}
        x^0 &:= (\phi^{-1})^0(y) = \frac{\sqrt{|\eta^{p+1,q+1}(z,z)|}}{|z^{d+1} + z^0|} >0, \\
        x^{1:d} &:= (\phi^{-1})^{1:d}(y) =  \frac{z^{1:d}}{z^{d+1} + z^0}.
    \end{align}
\end{Thm}

\begin{Rem}
    \label{rem:para-ads}
    Note that in \Cref{thm:para-ads} in the Euclidean case, i.e.\ if $q=0$, the subspace $\Zcal$ is empty, as $z^0 + z^{d+1} = 0$ would imply:
    \begin{align}
        0 &> \eta^{p+1,q+1}(z,z) \\
        &= |z^0|^2 + \eta^{p,q}(z^{1:d},z^{1:d}) - |z^{d+1}|^2 \\
        &= (z^0+z^{d+1})(z^0-z^{d+1}) + \eta^{p,q}(z^{1:d},z^{1:d}) \\
        &= \eta^{p,q}(z^{1:d},z^{1:d}),
    \end{align}
    which is not possible for $q=0$.
    It follows that for $q=0$ we get isometries:
    \begin{align}
        \Abb^{p+1,0} \cong \tAdS^{p+1,0} &\cong \AdS^{p+1,0}.
    \end{align}
\end{Rem}

\begin{Rem}
 For $p,q \ge 0$, $d:=p+q$, consider the map:
 \begin{align}
     \bar \phi: \R_{\ge 0} \times \R^{p,q} &\to \Pbb^{d+1}, & x &\mapsto [1-\eta^{p+1,q}(x,x) : 2x^{1:d} : 1+\eta^{p+1,q}(x,x) ].
 \end{align}
 Then $\bar \phi$ is well-defined. 
 For $x^0=0$ the map $\bar \phi$ coincides with the conformal embedding $\iota$ of $\R^{p,q}$ into the conformal compactification $\Mbb^{p,q} \ins \Pbb^{d+1}$ of $\R^{p,q}$, and, for $x^0>0$, $\bar \phi$ coincides with the isometric embedding $\phi$ of $\Abb^{p+1,q}$ into $\AdS^{p+1,q} \ins \Pbb^{d+1}$.
\end{Rem}

\subsection{The Euclidean case ($p \ge 2$ and $q=0$)}

In the following we now consider the Euclidean case, i.e.\ $q=0$, for $p \ge 2$, $d=p$.

\begin{Rem}
    Note that we have:
    \begin{align}
        \Sbb^0 &= \lC \pm 1 \rC, & \Sbb^{d,0} &= \Sbb^d \times \lC \pm 1 \rC.
    \end{align}
    So, for $q=0$, This thus simplifies the description of the conformal compactification, the corresponding de Sitter and anti-de Sitter space.
\end{Rem}

\begin{Cor}[The conformal compactification of the Euclidean space]
    We have the isometry for $d \ge 0$ and $q=0$, $p=d$:
    \begin{align}
        \Mbb^{d,0} &\cong \Sbb^d \times \lC +1 \rC \cong \Sbb^d, \\
        [z] & \mapsto \lp \frac{z^{0:d}}{z^{d+1}}, +1 \rp \mapsto \frac{z^{0:d}}{z^{d+1}}.
    \end{align}
    The conformal embedding is then given as:
    \begin{align}
        \tau_+: \R^{d,0} & \to \Mbb^{d,0} \cong \Sbb^d \ins \R^{d+1}, & \tau_+(x) &:= \lp \frac{1-\eta^{d,0}(x,x)}{1+\eta^{d,0}(x,x)}, \frac{2x}{1+\eta^{d,0}(x,x)} \rp,
    \end{align}
    which coincides with the invers stereographic projection (from the ``south pole'').

    The group $\POr(d+1,1)$ then acts on $\Sbb^d$ as follows, $[\Lambda] \in \POr(d+1,1)$, $y=y^{0:d} \in \Sbb^d$:
    \begin{align}
        [\Lambda].y & = \frac{\lp\Lambda\matb{y\\1}\rp^{0:d}}{\lp\Lambda\matb{y\\1}\rp^{d+1}},
    \end{align}
    where $\Lambda\matb{y\\1}$ denotes the matrix vector product of $(d+2)\times(d+2)$-matrix $\Lambda$ and the vector $y$ with a $1$ appended as the last component. Note that the indices of the components of $y \in \Sbb^d \ins \R^{d+1}$ range from $i=0,1,\dots,d$.
\end{Cor}

\begin{Rem}
    Note that we always have:
    \begin{align}
        \Isom(\Sbb^d) &\cong \Or(d+1),
    \end{align}
    via the usual action
    and for $d \ge 2$:
    \begin{align}
        \ConfDiff(\Sbb^d) &\cong \POr(d+1,1),
    \end{align}
    via the action above. The latter is part of \Cref{thm:conf-grp}. 
\end{Rem}

\begin{Rem}
    Recall that by \Cref{prp:para-ads}, \Cref{thm:para-ads} and \Cref{rem:para-ads} for $q=0$ we have isometries:
    \begin{align}
        \Abb^{d+1,0} \cong \tAdS^{d+1,0} &\cong \AdS^{d+1,0},
    \end{align}
    where we can slighly re-write\footnote{Note that for $y \in \R^{d+1,1}$ with $\eta^{d+1,1}(y,y) < 0$ we have: $|y^{d+1}|^2 > |y^0|^2$. So, for those $y$, we have the equivalence: $y^{d+1} > 0 \iff y^0+y^{d+1} > 0$.} $\tAdS^{d+1,0}$ as:
    \begin{align}
        \tAdS^{d+1,0}  &= \lC y \in \R^{d+1,1} \st \eta^{d+1,1}(y,y) = -1, y^{d+1} > 0 \rC.
    \end{align}
    With this we get the isometry:
    \begin{align}
        \AdS^{d+1,0} & \cong \tAdS^{d+1,0} \ins \R^{d+1,1}, & [z] & \mapsto \frac{\sgn(z^{d+1})}{\sqrt{|\eta^{d+1,1}(z,z)|}}\cdot z =: y.
    \end{align}
    Note that then:
    \begin{align}
        [y] &= [z] \in \AdS^{d+1}  \ins \Pbb^{d+1}.
    \end{align}
    Furthermore, for $[\Lambda] \in \POr(d+1,1)$ we define the action on $y \in \tAdS^{d+1,0}$:
    \begin{align}
        [\Lambda].y &= \sgn((\Lambda y)^{d+1}) \cdot (\Lambda y),
    \end{align}
    where $\Lambda y$ denotes the usual matrix product.
    Note that this is well-defined and that: 
    \begin{align} 
        [\Lambda].y &\in \tAdS^{d+1,0}, & [[\Lambda].y] = [\Lambda y] = [\Lambda][y] \in \AdS^{d+1,0}. 
    \end{align}
    This shows that the above isometry $\tAdS^{d+1,0} \cong \AdS^{d+1,0}$ is also $\POr(d+1,1)$-equivariant.
\end{Rem}

\begin{DefLem}[Geodesic distance of $\AdS^{d+1,0}$]
    The \emph{geodesic distance} on $\tAdS^{d+1,0}$, $\AdS^{d+1,0}$ and $\Abb^{d+1,0}$, resp., is given by:
    \begin{align}
        d^{\tAdS^{d+1,0}}(y_1,y_2)&:= \arccosh\lp |\eta^{d+1,1}(y_1,y_2)| \rp, \\
        d^{\AdS^{d+1,0}}([z_1],[z_2])&:= \arccosh\lp \frac{|\eta^{d+1,1}(z_1,z_2)|}{\sqrt{|\eta^{d+1,1}(z_1,z_1)|\cdot|\eta^{d+1,1}(z_2,z_2)|}} \rp, \\
        d^{\Abb^{d+1,0}}(x_1,x_2)&:= \arccosh\lp \frac{|x_1^0|^2 + |x_2^0|^2 + \eta^{d,0}(x_1^{1:d},x_2^{1:d})}{2x_1^0x_2^0} \rp.
    \end{align}
\end{DefLem}